\documentclass[runningheads]{llncs}
\usepackage{graphicx}
\usepackage{comment}
\usepackage{amsmath,amssymb} 
\usepackage{color}
\usepackage{capt-of}
\usepackage{wrapfig}
\usepackage{hyperref} 
\usepackage[mathscr]{euscript}
\DeclareSymbolFont{rsfs}{U}{rsfs}{m}{n}
\DeclareSymbolFontAlphabet{\mathscrsfs}{rsfs}

\newcommand{\secref}[1]{Section~\ref{#1}}
\newcommand{\figref}[1]{Figure~\ref{#1}}
\newcommand{\defeq}{\raisebox{-0.15\totalheight}{$\triangleq$}}
\newcommand{\informationloss}{\hspace{1mm}\raisebox{-0.30\totalheight}{\includegraphics[width=0.1in]{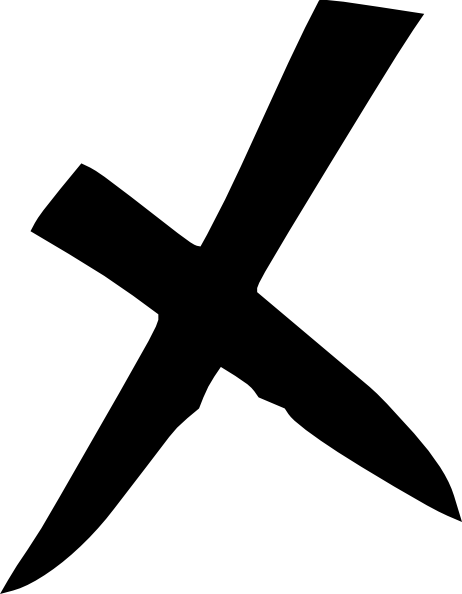}} \scriptsize{ (Info Loss)}}
\newcommand{\overfitting}{\hspace{1mm}\raisebox{-0.30\totalheight}{\includegraphics[width=0.1in]{arxiv_figures/x.png}} \scriptsize{ (Overfit)}}
\newcommand{\forgetting}{\hspace{1mm}\raisebox{-0.30\totalheight}{\includegraphics[width=0.1in]{arxiv_figures/x.png}} \scriptsize{ (Forgetting)}}
\newcommand{\prettycheck}{\hspace{1mm}\raisebox{-0.20\totalheight}{\includegraphics[width=0.1in]{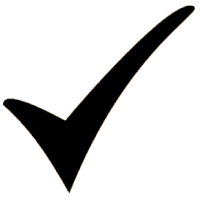}}}

\newcommand{\sidetuning}{\emph{side-tuning}}
\newcommand{\Sidetuning}{\emph{Side-tuning}}
\newcommand{\sidetune}{\emph{side-tune}}
\newcommand{\Sidetune}{\emph{Side-tune}}
\newcommand{\itaskonomy}{iTaskonomy}

\begin{document}
\pagestyle{headings}
\mainmatter
\def\ECCVSubNumber{1104}
\title{Side-Tuning: A Baseline for Network Adaptation via Additive Side Networks}
\titlerunning{Side-Tuning: A Baseline for Network Adaptation via Additive Side Networks}
\author{
    Jeffrey O. Zhang \inst{1} \and
    Alexander Sax \inst{1} \and
    Amir Zamir \inst{3} \and
    Leonidas Guibas \inst{2} \and
    Jitendra Malik \inst{1}
}
\authorrunning{Zhang et al.}
\institute{UC Berkeley \and Stanford University \and Swiss Federal Institute of Technology (EPFL) \\ 	\vspace{2mm}\href{http://sidetuning.berkeley.edu}{http://sidetuning.berkeley.edu\vspace{-9pt}}}

\maketitle

\begin{abstract}
When training a neural network for a desired task, one may prefer to adapt a \textbf{pre-trained} network rather than starting from randomly initialized weights. Adaptation can be useful in cases when training data is scarce, when a single learner needs to perform multiple tasks, or when one wishes to encode priors in the network. The most commonly employed approaches for network adaptation are \textbf{fine-tuning} and using the pre-trained network as a fixed \textbf{feature extractor}, among others. 

In this paper, we propose a straightforward alternative: \textbf{side-tuning}. Side-tuning adapts a pre-trained network by training a lightweight ``side" network that is fused with the (unchanged) pre-trained network via summation. This simple method works as well as or better than existing solutions and it resolves some of the basic issues with fine-tuning, fixed features, and other common approaches. In particular, side-tuning is less prone to overfitting, is asymptotically consistent, and does not suffer from catastrophic forgetting in incremental learning.  We demonstrate the performance of side-tuning under a diverse set of scenarios, including incremental learning (iCIFAR, \itaskonomy{}), reinforcement learning, imitation learning (visual navigation in Habitat), NLP question-answering (SQuAD v2), and single-task transfer learning (Taskonomy), with consistently promising results.
\end{abstract}

\section{Introduction}
\begin{figure}
    \vspace{-5mm}
    \centering
    \hspace{0mm}
    \includegraphics[width=0.6\columnwidth]{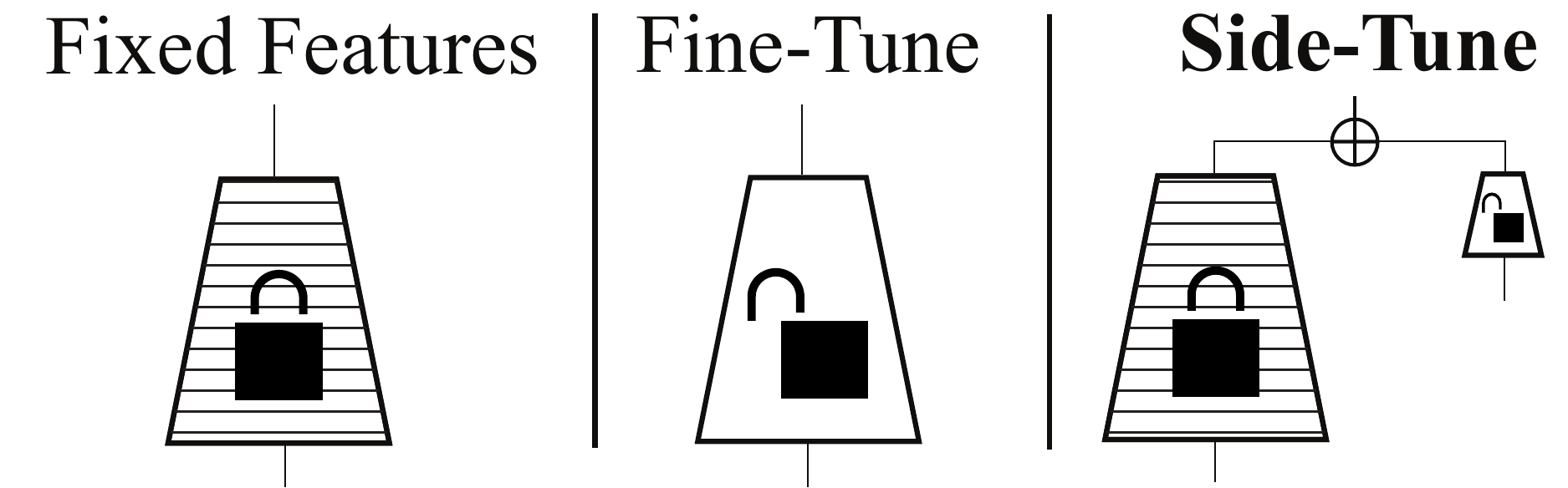}
    \vspace{0mm}
    \caption{\small{\textbf{The \emph{side-tuning} framework vs. common alternatives \emph{fine-tuning} and \emph{fixed features}}}. Given a pre-trained network that should be adapted to a new task, \emph{fine-tuning} re-trains the weights in the pretrained network and \emph{fixed feature extraction} trains a readout function with no re-training of the pre-trained weights. In contrast, \emph{Side-tuning} adapts the pre-trained network by training a lightweight conditioned ``side" network that is fused with the (unchanged) pre-trained network using summation.   
    }
    \vspace{-4mm}
    \label{fig:pull}
\end{figure}

 The goal of \sidetuning{} (and generally network adaptation) is to capitalize on a pretrained model to better learn one or more novel tasks.  
 The \sidetuning{} approach is straightforward: it assumes access to a given (base) model $B: \mathbb{X} \rightarrow \mathbb{Y}$ that maps the input $x$ onto some representation $y$. 
 \Sidetuning{} then \emph{learns} a side model $S: \mathbb{X} \rightarrow \mathbb{Y}$, so that the curated representations for the target task are 
\vspace{-0pt}\[R(x)~ \defeq~B(x) \oplus S(x),\]
for some combining operation $\oplus$. For example, choosing $B(x) \oplus S(x)~\defeq~\alpha B(x) + (1 - \alpha) S(x)$ (commonly called $\alpha$-blending) reduces the side-tuning approach to: fine-tuning, feature extraction, and stage-wise training, depending on $\alpha$ (Fig.~\ref{fig:methodology}, right). Hence those can be viewed as special cases of the side-tuning approach (\figref{fig:pull}). 


Side-tuning is an example of an \emph{additive learning} approach, one that adds new parameters for each new task.
Since \sidetuning{} does not change the base model, it, by design, adapts to a target task without degrading performance on the base task. Unlike many other additive approaches, \sidetuning{} places no constraints on the structure of the base model or side network, allowing for the architecture and sizes to vary independently. In particular, while other approaches require the side network to scale with the base network, \sidetuning{} can use tiny networks when the base only requires minor updates. By adding fewer parameters per task, \sidetuning{} can learn more tasks before the model grows large enough to require parameter consolidation. 

Substitutive methods instead opt for a single large model that is updated on each task. These methods often require adding additional constraints per-task in order to prevent inter-task interference~\cite{kirkpatrick2017overcoming,schwarz2018progress}. \Sidetuning{} does not require such regularization since the base remains untouched.

Compared to existing state-of-the-art network adaptation and incremental learning\footnote{Also referred to as lifelong or continual learning.} approaches, we find that the more complex methods perform no better---and often worse---than \sidetuning{}.

This straightforward mechanism deals with the key challenges of incremental learning (Sec. \ref{sec:forgetting_exps}). Namely, it does not suffer from either:
\begin{itemize}
    \item \textbf{Catastrophic forgetting}: tendency of a network to lose previously learned knowledge upon learning new information.
    \item \textbf{Rigidity}: Increasing inability of a network to adapt to new problems as it accrues constraints from previous problems. Note: Incremental learning literature sometimes calls this \emph{intransigence} \cite{chaudhry2018riemannian}. We prefer \emph{rigidity} as it is clear and widely used in psychology, dating back over 70 years~\cite{wiki:rigidity,leach1967critical}. 
\end{itemize}

We test \sidetuning{} for incremental learning on iCIFAR and the more challenging iTaskonomy dataset, which we introduce, finding that incremental learning methods that work on iCIFAR often do not work as well in the more demanding setup. On these datasets, \sidetuning{} uses side networks that are much smaller than the base. Consequently, even without consolidation, \sidetuning{} uses fewer learnable parameters than the alternative methods. 

Finally, because \sidetuning{} treats the base model as a black-box, it can be used with non-network sources of information such as a decision trees or oracle information on a related task (see Section~\ref{sec:learning_mechanics}). Thus, \sidetuning{} can be applied even when other model adaptation techniques cannot.

\begin{table}
    \vspace{-3mm}
    \centering
        \footnotesize
		\begin{tabular}{c|cc|c|}
			\hline

			\multicolumn{1}{|c|}{}& \multicolumn{2}{c|}{\textbf{1 Target Task}} & \multicolumn{1}{c|}{\textbf{$>$ 1 Target Tasks}} \\
			
			\multicolumn{1}{|c|}{Method} & \multicolumn{1}{c}{Low Data} & \multicolumn{1}{c|}{High Data} & \multicolumn{1}{c|}{(incremental)} \\
			\hline
			
			\hline
			\multicolumn{1}{|c|}{{Fixed features}} & \multicolumn{1}{l}{\prettycheck} & \multicolumn{1}{l|}{\informationloss}  & \multicolumn{1}{l|}{\informationloss} \\
			\multicolumn{1}{|c|}{{Fine-tuning}} & \multicolumn{1}{l}{\overfitting} & \multicolumn{1}{l|}{\prettycheck}  & \multicolumn{1}{l|}{\forgetting} \\

	    	\hline
        	\multicolumn{1}{|c|}{\textbf{\Sidetuning{}}} & \multicolumn{1}{l}{\prettycheck} & \multicolumn{1}{l|}{\prettycheck}  & \multicolumn{1}{l|}{\prettycheck} \\
	    	
	    	\hline
    \end{tabular} \\
    \vspace{2mm} 
    \caption{\footnotesize{\textbf{Advantages of \sidetuning{} vs. representative alternatives}. Fixed features provide a fixed representation and if the pretrained model discarded important information then there is no way to recover the lost details. This often leads to modest performance. On the other hand, fine-tuning has a large number of learnable parameters which leads to overfitting. \Sidetuning{} is a simple method that addresses these limitations with a small number of learnable parameters.}}
    \label{tab:adv}
    \vspace{-15mm}
\end{table}

\section{Related Work}
\label{sec:related_work}

Broadly speaking, network adaptation methods either overwrite existing parameters (\emph{substitutive} methods) or freeze them and add new parameters (\emph{additive} learning).

\textbf{Substitutive Methods} modify an existing network to solve a new task by updating some or all of the network weights (simplest approach being fine-tuning). A large body of constraint-based methods focuses on how to regularize these updates in order to prevent forgetting earlier tasks. Methods such as \cite{schwarz2018progress,kirkpatrick2017overcoming,li2017learning} impose additional constraints for each new task, which slows down learning on later tasks (see Sec. \ref{sec:rigidity} on rigidity, \cite{ChaudhryAGEM}). Other methods such as \cite{cheung2019superposition} relegate each task to approximately orthogonal subspaces but are then unable to transfer information across tasks. \Sidetuning{} does not require such regularization since the base remains untouched.

\textbf{Additive Methods} methods circumvent forgetting by freezing the weights and adding a small number of new parameters per task. One economical approach is to use off-the-shelf-features with one or more readout layers~\cite{sharif2014cnn}. However, off-the-shelf features cannot be updated for the new task, and so recent work has focused on how features can be modulated by applying per-task learned weight masks \cite{mallya2018piggyback,rosenfeld2018incremental}, by pruning~\cite{mallya2018packnet}, or by hard attention~\cite{serra2018overcoming}.

If information is missing from the original features, then recovering that information might require adding additional weights. Works such as \cite{prognets16,lee2020residual} introduce a new network with independent access to the input and connect to various layers from the original network. Other works like~\cite{rebuffi2017learning,bilen2017universal,rebuffi2018efficient} learn task-dependent parameters (e.g. separate batch norm, linear layers) that are inserted directly into the existing network. Tying the new weights directly into the original network architecture often requires making restrictive assumptions about the original network architecture (e.g. that it must be a ResNet~\cite{he2016deep}).

Unlike these previous works, \sidetuning{} uses only late fusion and makes no assumptions about the base network. This means it can be applied on a larger class of models. While simpler, the results suggest that \sidetuning{} offers similar or better performance to the more complex alternatives and calls into question whether that complexity buys much in practice.

\textbf{Residual Learning} exploits the fact that it is often easier to approximate a difference rather than the original function. This has been successfully used in ResNets \cite{he2016deep} where a residual is learned on top of the identity function prior. Some network adaptation methods insert new residual-modeling parameters directly into the base architecture~\cite{lee2020residual,rebuffi2017learning}. Residual learning has also been explored in robotics as residual RL \cite{johannink2019residual_levine,residualRLKaelbling}, in which we train an agent for a single task by first taking a coarse policy (e.g. behavior cloning) and then training a residual network on top (using RL). For a single task, iteratively learning residuals is known as gradient boosting, but \sidetuning{} adds a side network to adapt a base representation for a \emph{new} task. We discuss the relationship in Sec.~\ref{sec:learning_mechanics}.

\textbf{Meta-learning}, unlike network adaptation approaches, seeks to create networks that are inherently adaptable. Typically this proceeds by training on tasks sampled from a known task distribution. \Sidetuning{} is fundamentally compatible with this formulation and with existing approaches (e.g.~\cite{FinnAL17}). Recent work suggests that these approaches work primarily by feature adaptation rather than rapid learning~\cite{Raghu2019}, and feature adaptation is also the motivation for our method.

\section{Side-Tuning: The Simplest Additive Approach}
\label{sec:framework}

\Sidetuning{} learns a side model $S(x)$ and combines it with a pre-trained base model $B(x)$. The representation for the target task is
$R(x)~ \defeq~B(x) \oplus S(x)$. 

\begin{figure*}
    \centering
    \includegraphics[width=0.8\columnwidth]{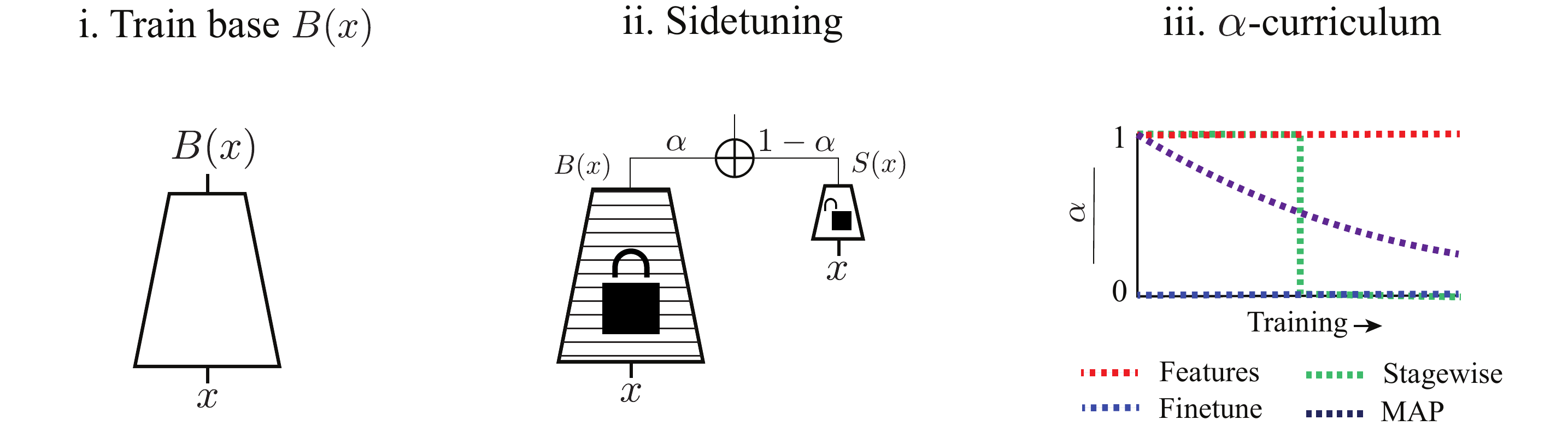}
    \vspace{-5mm}
    \caption{\footnotesize{\textbf{Mechanics of \sidetuning{}.} (i) \Sidetuning{} takes some core network and adapts it to a new task by training a side network. (ii) Connectivity structure when using \sidetuning{} along with $\alpha$-blending. (iii) Some of the existing common adaptation methods turn out to be special cases of an alpha blending with a side network. In particular: fine-tuning, feature extraction, and other approaches are \sidetuning{} with a fixed curriculum on the blending parameter $\alpha$, as shown in the plot.}}
    \label{fig:methodology}
\end{figure*}

\subsection{Architectural Elements}

\textbf{Base Model.}
The base model $B(x)$ provides some core cognition or perception, and we put no restrictions on how $B(x)$ is computed. We never update $B(x)$, and in our approach it has zero learnable parameters.  $B(x)$ could be a decision tree or an oracle for another task (experiments with this setup shown in \secref{sec:learning_mechanics}). We consider several choices for $B(x)$ in Section~\ref{sec:learning_mechanics}, but the simplest choice is just a pretrained network. 
\vspace{1mm}

\textbf{Side Model.}
Unlike the base model, the side network, $S(x)$, is updated during training; learning a residual that we apply on top the base representation. One crucial component of the framework is that the complexity of the side network can scale to the difficulty of the problem at hand. When the base is relevant and requires only a minor update, a very simple side network can suffice. 

Since the side network’s role is to amend the base network to a new task, we initialize the side network as a copy of the base. When the forms of the base and side networks differ, we initialize the side network through knowledge distillation~\cite{hinton2015distilling}. We investigate side network design decisions in Sec.~\ref{sec:learning_mechanics}. In general, we found \sidetuning{} to perform well in a variety of settings and setups.

\vspace{1mm}

\textbf{Combining Base and Side Representations.}
\label{sec:alphablending}

\Sidetuning{} admits many options for the combination operator, $\oplus$, and we compare several in Section~\ref{sec:merge_methods}. We observe that alpha blending, $B(x) \oplus S(x)~\defeq~\alpha B(x) + (1 - \alpha) S(x)$, where $\alpha$ is treated as a learnable parameter works well and $\alpha$ correlates with task relevance (see Section~\ref{sec:learning_mechanics}). 

While simple, alpha blending is expressive enough that it encompasses several common transfer learning approaches. As shown in Figure~\ref{fig:methodology}(iii), \sidetuning{} is equivalent to feature extraction when $\alpha=1$. When $\alpha=0$, \sidetuning{} is instead equivalent to fine-tuning if the side network has the same architecture the base. If we allow $\alpha$ to vary during training, then switching $\alpha$ from 1 to 0 is equivalent to the common (stage-wise) training curriculum in RL where a policy is trained on top of some fixed features that are unlocked partway through training.

When minimizing estimation error there is often a tradeoff between the bias and variance contributions~\cite{geman1992neural} (see Table~\ref{tab:adv}). Feature extraction locks the weights and corresponds to a point-mass prior while fine-tuning is an uninformative prior yielding a low-bias high-variance estimator. \Sidetuning{} aims to leverage the (useful) bias from those original features while making the representation \emph{asymptotically consistent} through updates to the residual side-network~\footnote{Sidetuning is one way of making features obey Cromwell's rule: ``I beseech you, in the bowels of Christ, think it possible that you may be mistaken.''}.

Given the bias-variance interpretation, a notable curriculum for $\alpha$ during training is $\alpha(N) = \frac{k}{k+N}$ for $k > 0$ (hyperbolic decay) where N is the number of training epochs.
This curriculum, placing less weight on the prior as more evidence accumulates, is suggestive of a \emph{maximum a posteriori} estimate and, like the MAP estimate, it converges to the MLE (fine-tuning).

\subsection{Side-Tuning for Incremental Learning}

We often care about the performance not only on the current target task but also on the previously learned tasks. This is the case for \emph{incremental learning}, where we want an agent that can learn a sequence of tasks $\mathcal{T}_1, ..., \mathcal{T}_m$ and is capable of reasonable performance across the entire set at the end of training. In this paradigm, catastrophic forgetting (diminished performance on $\{\mathcal{T}_1, ..., \mathcal{T}_{m-1}\}$ due to learning $\mathcal{T}_m$) becomes a major issue.

\setlength{\columnsep}{7pt}

\begin{wrapfigure}{r}{0.45\textwidth}
    \vspace{-6mm}
    \centering
    \includegraphics[width=0.4\columnwidth]{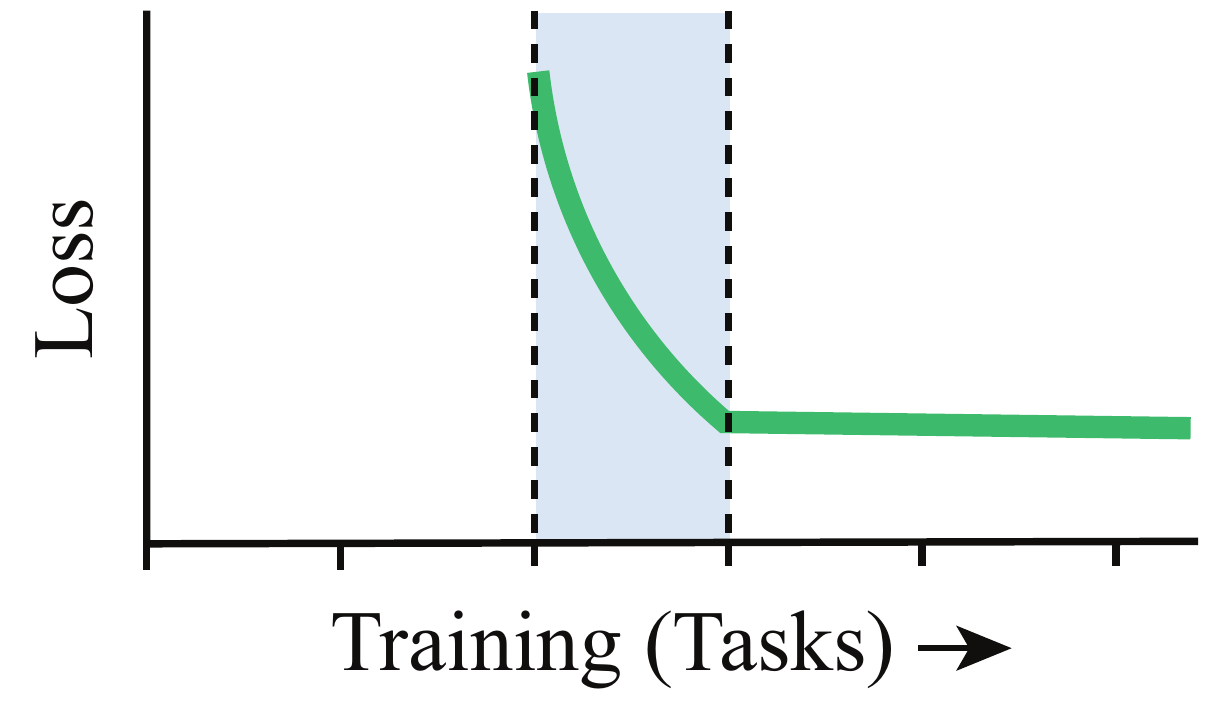}
    \vspace{-4mm}
    \caption{\footnotesize{\textbf{Theoretical learning curve of \sidetuning{}.} The model learns during task-specific training and those weights are subsequently frozen, preserving performance.}}
    \vspace{-6mm}
    \label{fig:tlc}
\end{wrapfigure}

In our experiments, we dedicate one new side network to each task. We define a task $\mathcal{T}: x \mapsto P(Y)$ as a mapping from inputs, $x$, to a probability distribution over the output space, $Y$. 
For example, $x$ is an RGB image mapped to probabilities over object classes, $Y$.
Datasets for a task are a set of pairs $\{(x,y)~|~y \sim \mathcal{T}(x)\}$. For task $\mathcal{T}_t$, our loss function is \vspace{-2mm} 

$$L(x_t,y_t) = \lVert D_t(\alpha_t B(x_t) + (1 - \alpha_t) S_t(x_t)) - y_t \rVert$$

where $t$ is the task number and $D_t$ is some decoder readout of the \sidetuning{} representation. 
This simple approach leads to the training curve in Figure~\ref{fig:tlc} with no possible catastrophic forgetting. Furthermore, since \sidetuning{} is independent of task order, training does not slow down as training progresses.
We observe that this approach provides a strong baseline for incremental learning, outperforming existing approaches in the literature while using fewer parameters on more tasks (in Section \ref{sec:lifelong}). 

\begin{figure*}[h]
    \centering
    \hspace{-1mm}\includegraphics[width=0.85\textwidth]{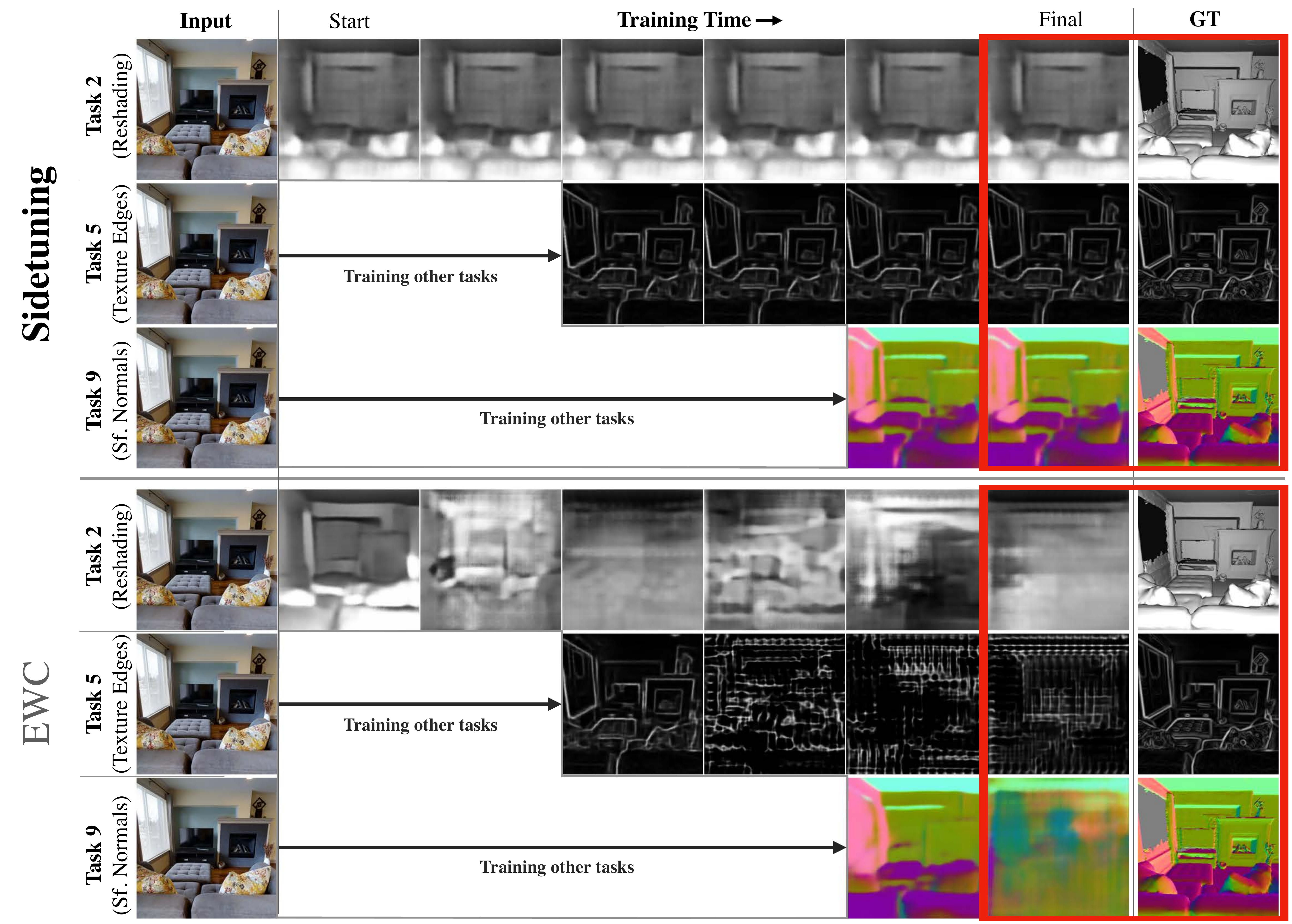}
    \caption{\footnotesize{\textbf{\Sidetuning{} does not forget in incremental learning.} Qualitative results for \itaskonomy{} with additive learning (\sidetuning{}, top 3 rows) and constraint-based substitutive learning (EWC, bottom 3 rows). Each row contains results for one task and columns show how predictions change over the course of training. Predictions from EWC quickly degrade over time, showing that EWC catastrophically forgets. Predictions from \sidetuning{} do not degrade, and the initial quality is better in later tasks (e.g. compare the table in surface normals). We provide additional comparisons (including for PSP, PNN) in the supplementary.}}
    \label{fig:lll_qual}
    \vspace{-4mm}
\end{figure*}

\Sidetuning{} naturally handles other continuous learning scenarios besides incremental learning. A related problem is that of continuous adaptation, where the agent needs to perform well (e.g. minimizing regret) on a stream of tasks with undefined boundaries and where there might very little data per task and no task repeats. 
As we show in Section~\ref{sec:lifelong}, inflexibility becomes a serious problem for constraint-based methods and task-specific performance declines after learning more than a handful of tasks. Moreover, continuous adaptation requires an online method as task boundaries must be detected and data cannot be replayed (e.g. to generate constraints for EWC). 

\Sidetuning{} could be applied to continuous adaptation by keeping a small working memory of cheap side networks that constantly adapt the base network to the input task. These side networks are small, easy to train, and when one of the networks begins performing poorly (e.g. signaling a distribution shift) that network can simply be discarded. This is an online approach, and online adaptation with small cheap networks has found recent success (e.g. in~\cite{RamananOnlineDistillation}).



\section{Experiments}
\label{others}


In the first section we show that when applied to the incremental learning setup, \sidetuning{} compares favorably to existing approaches on both iCIFAR and the more challenging \itaskonomy{} dataset. We then extend this to multiple domains (computer vision, RL, imitation learning, NLP) in the simplified scenario for $m=2$ tasks (transfer learning). Finally, we interpret \sidetuning{} in a series of analysis experiments.


\subsection{Baselines}
We provide comparisons of \sidetuning{} against the following methods:

\begin{description}
\begin{footnotesize}
    \item \textbf{Scratch}: The network is initialized with appropriate random weights and trained using minibatch SGD with Adam \cite{kingma2014adam}.
    \item \textbf{Feature extraction (features)}: The pretrained base network is used as-is and is not updated during training.
    \item \textbf{Fine-tuning}: An umbrella term that encompasses a variety of techniques, we consider a more narrow definition where pretrained weights are used as initialization and then training proceeds as in \emph{scratch}.
    \item \textbf{Elastic Weight Consolidation (EWC).} A constraint-based substitutive approach from~\cite{kirkpatrick2017overcoming}. We use the formulation from~\cite{schwarz2018progress} which scales better.
    \item \textbf{Parameter Superposition (PSP)}: A parameter-masking substitutive approach from~\cite{cheung2019superposition} that attempts to make tasks independent from one another by mapping the weights to approximately orthogonal spaces. 
    \item \textbf{Progressive Neural Network (PNN)}: An additive approach from~\cite{prognets16} which utilizes many lateral connections between the base and side networks. Requires the architecture of the base and side networks to be the same or similar.
    \item \textbf{Piggyback (PB)}: Learns task-dependent binary weight masks~\cite{mallya2018piggyback}. 
    \item \textbf{Residual Adapters (RA)}: An additive approach which learns task-dependent batch-norm and linear layers between layers in an existing network~\cite{rebuffi2017learning,rebuffi2018efficient}.
    \item \textbf{Independent}: Each task uses a pretrained network trained independently for that task. This method uses far more learnable parameters than all the alternatives (e.g. saving a separate ResNet-50 for each task) and achieves strong performance.
\end{footnotesize}
\end{description}

\subsection{Incremental Learning}
\label{sec:lifelong}

On both the incremental Taskonomy~\cite{taskonomy2018} (\itaskonomy{}) and incremental CIFAR (iCIFAR~\cite{icifar}) datasets, \sidetuning{} performs competitively against existing incremental learning approaches while using fewer parameters\footnote{Full experimental details (e.g. architecture) provided in the supplementary.}. On the more challenging Taskonomy dataset, it outperforms other approaches.

\begin{itemize}
    \item \textbf{iCIFAR.} Comprises 10 subsequent tasks by partitioning CIFAR-100~\cite{icifar} into 10 disjoint sets of 10-classes each. Images are $32{\times}32$ RGB. First, we pretrain the base network (ResNet-44) on CIFAR-10. We then train on each subtask for 20k steps before moving to the next one. The SotA substitutive baselines (EWC and PSP) update the base network for each task (683K parameters), while \sidetuning{} updates a four layer convolutional network per task (259K parameters after 10 tasks).
    \item \textbf{\itaskonomy{}.} Taskonomy~\cite{taskonomy2018} is significantly more challenging than CIFAR-100 and includes multiple computer vision tasks beyond object classification: including 2D (e.g. edge detection), 3D  (e.g. surface normal estimation), and semantic (e.g. object classification) tasks. We note that approaches which work well on iCIFAR often do quite poorly in the more realistic setting. We created iTaskonomy by selecting all (12) tasks that make predictions from a single RGB image, and then created an incremental learning setup by selecting a random order in which to learn these tasks (starting with \emph{curvature}). The images are $256{\times}256$ and we use a ResNet-50 for the base network and a 5-layer convolutional network for the \sidetuning{} side network. The number of learnable network parameters used across all tasks is 24.6M for EWC and PSP, and 11.0M for \sidetuning{}\footnote{Numbers not counting readout parameters, which are common between all methods.}. 
\end{itemize}

\begin{figure*}
    \vspace{-4mm}
    \scriptsize
    \centering
    \begin{minipage}{0.48\textwidth}
        \centering
        \includegraphics[width=1\textwidth]{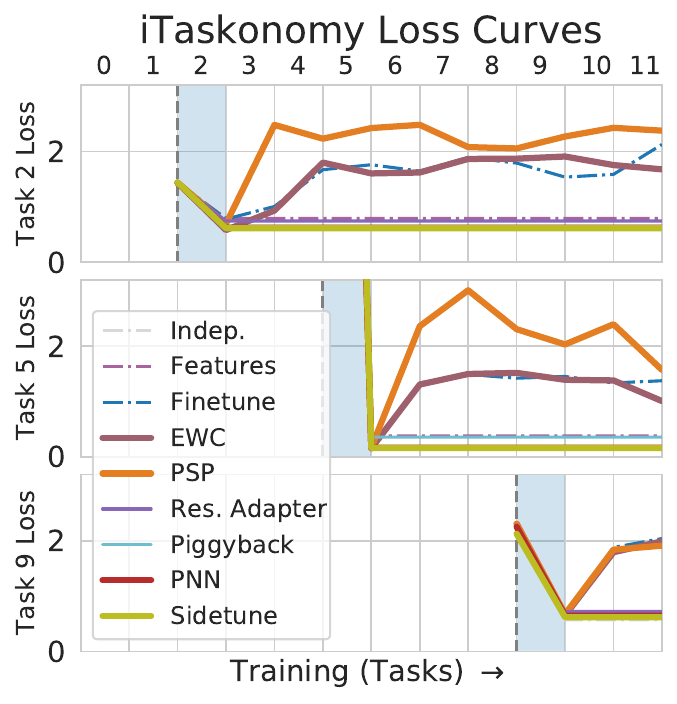}
        \vspace{0mm}
    \end{minipage}    
    \begin{minipage}{0.48\textwidth}
        \centering
        \includegraphics[width=1.0\textwidth]{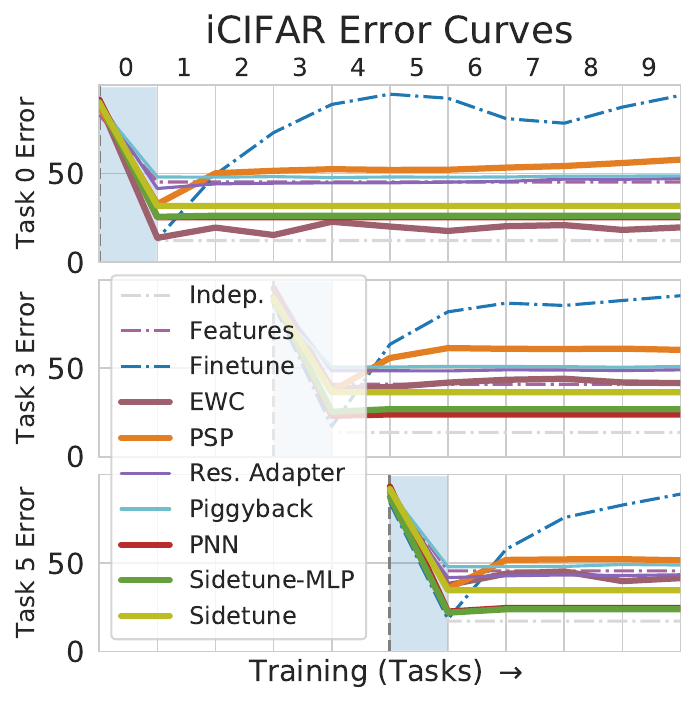}
        \vspace{0mm}
    \end{minipage}
    \vspace{-7mm}
    \caption{\footnotesize{\textbf{Incremental Learning on \itaskonomy{} and iCIFAR}. The above curves show loss and error for three tasks on \itaskonomy{} (left) and iCIFAR (right) datasets. The fact that \sidetuning{} losses are flat after training (as we go right) shows that it does not forget previously learned tasks. That performance remains consistent even on later tasks (as we go down), showing that \sidetuning{} does not become rigid. Substitutive methods show clear forgetting (e.g. PSP) and/or rigidity (e.g. EWC).} In \itaskonomy{}, PNN and Independent are hidden under Sidetune.
    }
    \label{fig:lll_quant}
    \vspace{-5mm}
\end{figure*}


\subsubsection{Catastrophic Forgetting}
\label{sec:forgetting_exps}
As expected, there is no catastrophic forgetting in \sidetuning{} and other additive methods. Figure~\ref{fig:lll_quant} shows that the error for \sidetuning{} does not increase after training (blue shaded region), while it increases sharply for the substitutive methods on both \itaskonomy{} and iCIFAR.

The difference is meaningful, and Figure~\ref{fig:lll_qual} shows sample predictions from \sidetuning{} and EWC for a few tasks during and after training. As is evident from the bottom rows, EWC exhibits catastrophic forgetting on all tasks (worse image quality as we move right). In contrast, \sidetuning{} (top) shows no forgetting and the final predictions are significantly closer to the ground-truth (boxed red). 

\subsubsection{Rigidity}\label{sec:rigidity}
\Sidetuning{} learns later tasks as easily as the first, while constraint-based methods such as EWC stagnate. The predictions for later tasks are significantly better using \sidetuning{} \textbf{even immediately after training and before any forgetting can occur} (e.g., \emph{surface normals} in Fig.~\ref{fig:lll_qual}).  

\begin{figure}[h]
    \scriptsize
    \centering
    \begin{minipage}{0.4\columnwidth}
        \centering
        \vspace{-1mm}
        \includegraphics[width=\columnwidth]{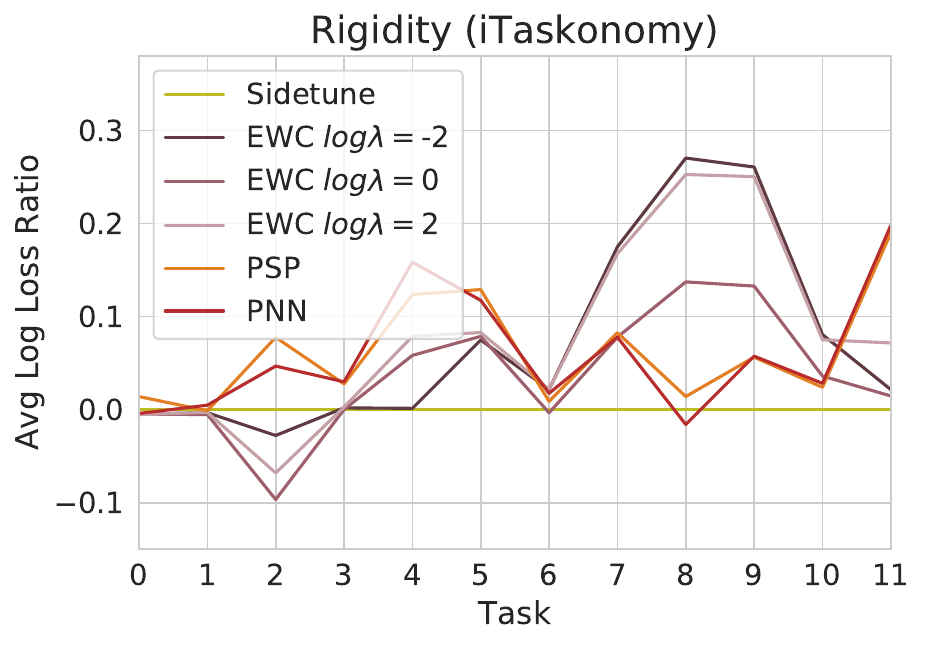}
        \vspace{-9mm}
    \end{minipage}
    \begin{minipage}{0.4\columnwidth}
        \centering
        \includegraphics[width=\columnwidth]{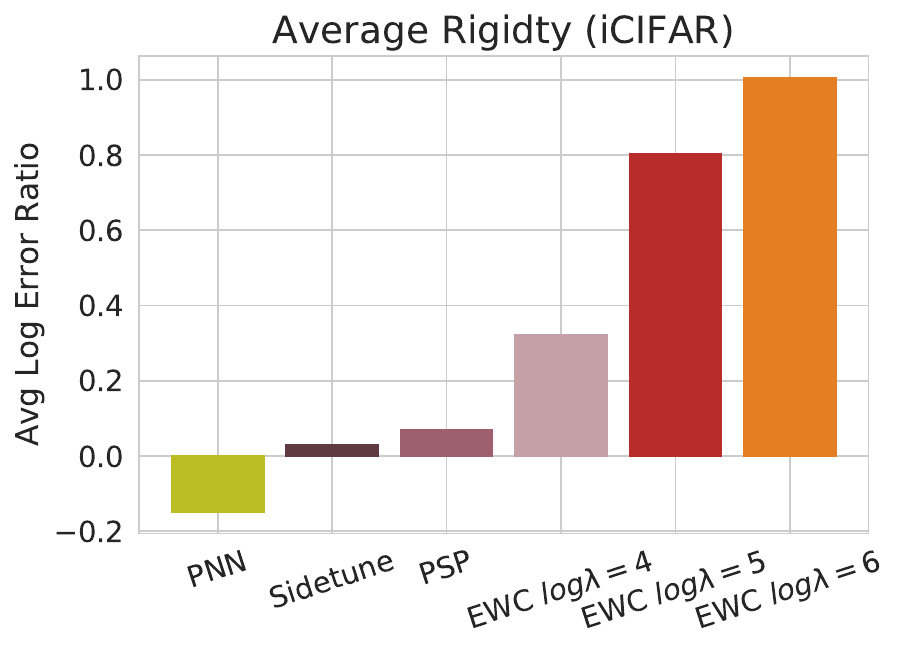}
        \vspace{-8mm}
    \end{minipage}
    \vspace{2mm}
    \caption{\footnotesize{\textbf{Rigidity (Intransigence) on \itaskonomy{} and iCIFAR}.  \Sidetuning{} always learns new tasks easily; EWC becomes increasingly unable to learn new tasks as training progresses. The same trend holds on iCIFAR (right).
    }}
\label{fig:rigidity}
\end{figure}
Figure~\ref{fig:rigidity} quantifies this slowdown. We measure rigidity as the log-ratio of the \emph{actual loss} of the $i$th task over the loss when that task is \emph{instead trained first} in the sequence. As expected, \sidetuning{} experiences effectively zero slowdown on both datasets. For EWC, the added constraints make learning new tasks increasingly difficult and rigidity increases with the number of tasks (Fig.~\ref{fig:rigidity}, left). PNN shows some positive transfer in iCIFAR (negative ratio value), but becomes rigid on the more challenging \itaskonomy{}, where tasks are more diverse.

\subsubsection{Final Performance}\label{sec:lll_perf}
Overall, \sidetuning{} significantly outperforms the substitutive methods while using fewer than half the number of trainable parameters. It is comparable with additive methods while remaining remarkably simpler. On iCIFAR, vanilla \sidetuning{} achieves a strong average rank (2.20 of 6, see Table~\ref{fig:rank}) and, when using the same combining operator (MLP) as PNN, is as good as the best-performing model without the additional lateral connections (see Figure~\ref{fig:lll_quant} right). On \itaskonomy{}, vanilla \sidetuning{} achieves the best average rank (1.33 of 6, while the next best is 2.42 by PNN, see Table~\ref{fig:rank}).


\begin{wrapfigure}{r}{0.5\textwidth}
    \vspace{-2mm}
    \scriptsize
    \centering
    \begin{tabular}{l|ll}
			\hline
			\multicolumn{1}{|c|}{}& \multicolumn{2}{c|}{Avg. Rank ($\downarrow$)} \\ 
			\multicolumn{1}{|c|}{Method} & \multicolumn{1}{c}{\itaskonomy{}} & \multicolumn{1}{c|}{iCIFAR} \\
			
			\hline
			\hline
			\multicolumn{1}{|c|}{{EWC ($\lambda=10^0,10^5$)}} & \multicolumn{1}{c}{5.25} & \multicolumn{1}{c|}{2.70} \\

			\multicolumn{1}{|c|}{{PSP}} & \multicolumn{1}{c}{5.25} & \multicolumn{1}{c|}{5.60} \\
			
			\multicolumn{1}{|c|}{{Res. Adapter}} & \multicolumn{1}{c}{3.58} & \multicolumn{1}{c|}{4.40} \\

			\multicolumn{1}{|c|}{{Piggyback}} & \multicolumn{1}{c}{3.17} & \multicolumn{1}{c|}{5.00} \\

			\multicolumn{1}{|c|}{{PNN}} & \multicolumn{1}{c}{2.42} & \multicolumn{1}{c|}{\textbf{1.10}} \\
			
			\hline
			\multicolumn{1}{|c|}{\Sidetune{}} & \multicolumn{1}{c}{\textbf{1.33}} & \multicolumn{1}{c|}{2.20}  \\
			\hline
	\end{tabular}
    \vspace{2mm}
    \captionof{table}{\footnotesize{\textbf{Average rank on \itaskonomy{} and iCIFAR.} Despite being simpler than alternatives, \sidetuning{} generally achieved a better average rank than other approaches. The difference increases on the more challenging Tasknomy dataset, where \sidetuning{} significantly outperformed all tested alternatives.}}
    \vspace{-4mm}
    \label{fig:rank}
\end{wrapfigure}

This is a direct result of the fact (shown above) that \sidetuning{} does not suffer from catastrophic forgetting or rigidity (intransigence). It is not due to the fact that the sidetuning structure is specially designed for these types of image tasks; it is not (we show in Sec.~\ref{sec:universality} that it performs well on other domains). In fact, the much larger networks used in EWC and PSP \emph{should} achieve better performance on any single task. For example, EWC produces sharper images early on in training, before it has had a chance to accumulate too many constraints (e.g. \emph{reshading} in Figure~\ref{fig:lll_qual}). But this factor was outweighed by \emph{side-tuning's} immunity from the effects of catastrophic forgetting and compunding rigidity.

\subsection{Universality of the Experimental Trends}
\label{sec:universality} 

In order to address the possibility that \sidetuning{} is somehow domain- or task-specific, we provide results showing that it is well-behaved in other settings. As the concern with \emph{additive learning} is mainly that it is too inflexible to learn new tasks, we compare with fine-tuning (which outperforms other incremental learning tasks when forgetting is not an issue). For extremely limited amounts of data, feature extraction can outperform fine-tuning. We show that \sidetuning{} generally performs as well as features or fine-tuning--whichever is better.

\begin{table*}
\vspace{-0mm}
\centering
\hspace{-15mm}
\begin{minipage}{0.15\textwidth}
    \centering
    \vspace{2.6mm}
    \setlength{\tabcolsep}{4pt}
    \scriptsize
    \centering
    \textbf{}
    \begin{tabular}{|l|}
			\hline
			\multicolumn{1}{|c|}{} \\ 
			\multicolumn{1}{|c|}{{Method}} \\
			
			\hline
			\hline
			\multicolumn{1}{|c|}{{Fine-tune}} \\

			\multicolumn{1}{|c|}{{Features}} \\

			\multicolumn{1}{|c|}{{Scratch}} \\
			
			\hline
			\multicolumn{1}{|c|}{\Sidetune{}} \\
            \hline
	\end{tabular} \\ 
\label{fig:method_names}
\end{minipage}
\hspace{-3mm}
\begin{minipage}{0.36\textwidth}
    \centering
    \setlength{\tabcolsep}{4pt}
    \scriptsize
    \centering
    \textbf{Transfer Learning in Taskonomy}
    \begin{tabular}{ll}
		\hline
		\multicolumn{2}{|c|}{From Curvature (100/4M ims.) } \\ 
		\multicolumn{1}{|c}{\tiny{Normals (MSE $\downarrow$)}} & \multicolumn{1}{c|}{\tiny{Obj. Cls. (Acc. $\uparrow$)}} \\
		
		\hline
		\hline
		\multicolumn{1}{|c}{\textbf{0.200 / 0.094}} & \multicolumn{1}{c|}{\textbf{24.6 / 62.8}} \\

		\multicolumn{1}{|c}{\textbf{0.204} / 0.117} & \multicolumn{1}{c|}{24.4 / 45.4} \\

		\multicolumn{1}{|c}{0.323 / \textbf{0.095}} & \multicolumn{1}{c|}{19.1 / 62.3} \\
		
		\hline
		\multicolumn{1}{|c}{\textbf{0.199 / 0.095}} & \multicolumn{1}{c|}{\textbf{24.8 / 63.3}}  \\
        \hline
	\end{tabular} \\ 
    \textbf{(a)}
\label{fig:transfer_vision}
\end{minipage}
\hspace{-3mm}
\begin{minipage}{0.16\textwidth}
    \centering
    \setlength{\tabcolsep}{4pt}
    \scriptsize
    \centering
    \textbf{QA on SQuAD}
    \begin{tabular}{|ll}
		\hline
		\multicolumn{2}{|c|}{Match ($\uparrow$)} \\ 
		\multicolumn{1}{|c}{\tiny{Exact}} & \multicolumn{1}{c|}{\tiny{F1}} \\
		\hline
		\hline
		\multicolumn{1}{|c}{\textbf{79.0}} & \multicolumn{1}{c|}{\textbf{82.2}} \\
		\multicolumn{1}{|c}{49.4} & \multicolumn{1}{c|}{49.5} \\
		\multicolumn{1}{|c}{0.98} & \multicolumn{1}{c|}{4.65} \\
		\hline
		\multicolumn{1}{|c}{\textbf{79.6}} & \multicolumn{1}{c|}{\textbf{82.7}}  \\
		\hline
	\end{tabular} \\
    \textbf{(b)}
    \label{fig:nlp_main}
\end{minipage} 
\hspace{-3mm}
\begin{minipage}{0.16\textwidth}
    \centering
    \setlength{\tabcolsep}{4pt}
    \scriptsize
    \centering
    \textbf{Navigation (IL)}
    \setlength{\tabcolsep}{2pt}
    \begin{tabular}{ll}
		\hline
		 \multicolumn{2}{|c|}{Nav. Rew. ($\uparrow$)} \\ 
		 \multicolumn{1}{|c}{\tiny{Curv.}} & \multicolumn{1}{c|}{\tiny{Denoise}} \\
		\hline
		\hline
		\multicolumn{1}{|c}{10.5} & \multicolumn{1}{c|}{9.2} \\
		 \multicolumn{1}{|c}{\textbf{11.2}} & \multicolumn{1}{c|}{8.2} \\
		\multicolumn{1}{|c}{9.4} & \multicolumn{1}{c|}{\textbf{9.4}} \\
		\hline
		\multicolumn{1}{|c}{\textbf{11.1}} & \multicolumn{1}{c|}{\textbf{9.5}}  \\
		\hline
	\end{tabular} \\
    \textbf{(c)}
	\label{fig:il_main}
\end{minipage}
\hspace{-3mm}
\begin{minipage}{0.17\textwidth}
    \centering
    \setlength{\tabcolsep}{2pt}
    \scriptsize
    \centering
    \textbf{Navigation (RL)}
    \scriptsize{
    \begin{tabular}{ll}
			\hline
			\multicolumn{2}{|c|}{Nav. Rew. ($\uparrow$)} \\ 
			\multicolumn{1}{|c}{\tiny{Curv.}} & \multicolumn{1}{c|}{\tiny{Denoise}} \\
			\hline
			\hline
            \multicolumn{1}{|c}{10.7} & \multicolumn{1}{c|}{\textbf{10.0}} \\
            \multicolumn{1}{|c}{\textbf{11.9}} & \multicolumn{1}{c|}{8.3} \\
			\multicolumn{1}{|c}{7.5} & \multicolumn{1}{c|}{7.5} \\
			\hline
			\multicolumn{1}{|c}{\textbf{11.8}} & \multicolumn{1}{c|}{\textbf{10.4}}  \\
			\hline
	\end{tabular} \\
	\textbf{(d)}
	\label{fig:rl}
	}
\end{minipage}
\hspace{-12mm}
\vspace{2mm}
\caption{\textbf{Side-tuning comparisons in other domains.} Sidetuning matched the adaptability of fine-tuning on large datasets, while performing as well or better than the best competing method in each domain:
\textbf{(a)} In Taskonomy, for \emph{Normal Estimation} or \emph{Object Classification} using a base trained for \emph{Curvatures} and either 100 or 4M images for transfer. Results using \emph{Obj. Cls.} base are similar and provided in the supplementary materials.
\textbf{(b)} In SQuAD v2 question-answering, using BERT instead of a convolutional architecture.
\textbf{(c)} In Habitat, learning to navigate by imitating expert navigation policies, using inputs based on either \emph{Curvature} or \emph{Denoising}. Finetuning does not perform as well in this domain.
\textbf{(d)} Using RL (PPO) and direct interaction instead of supervised learning for navigation.}
\vspace{-7mm}
\label{fig:universality}
\end{table*}

\textbf{Transfer learning in Taskonomy.} We trained networks to perform one of three target tasks (object classification, surface normal estimation, and curvature estimation) on the Taskonomy dataset~\cite{taskonomy2018} and varied the size of the training set $N \in \{100, 4\times 10^6\}$. In each scenario, the base network was trained (from scratch) to predict one of the non-target tasks. The side network was a copy of the original base network. We experimented with a version of fine-tuning that updated both the base and side networks; the results were similar to standard fine-tuning 
\footnote{\label{defer}We defer remaining experimental details (learning rate, full architecture, etc.) to the supplementary materials. See provided code for full details.}.
In all scenarios, \sidetuning{} successfully matched the adaptiveness of fine-tuning, and significantly outperformed learning from scratch, as shown in Table~\ref{fig:universality}a. The additional structure of the  frozen base did not constrain performance with large amounts of data (4M images), and \sidetuning{} performed as well as (and sometimes slightly better than) fine-tuning.



\textbf{Question-Answering in SQuAD v2.} We also evaluated \sidetuning{} on a question-answering task (SQuAD v2~\cite{RajpurkarSQuAD2}) using a non-convolutional architecture. We use a pretrained BERT~\cite{BERT} model for our base, and a second for the side network. Unlike in the previous experiments, BERT uses attention and no convolutions. Still, \sidetuning{} adapts to the new task just as well as fine-tuning, outperforming features and scratch (Table~\ref{fig:universality}b).

\textbf{Imitation Learning for Navigation in Habitat.}
We trained an agent to navigate to a target coordinate in the Habitat environment. The agent is provided with both RGB input image and also an occupancy map of previous locations. The map does not contain any information about the environment---just previous locations. In this section we use Behavior Cloning to train an agent to imitate experts following the shortest path on 49k trajectories in 72 buildings. The agents are evaluated in 14 held-out validation buildings. Depending on what the base network was trained on, the source task might be useful (\emph{Curvature}) or harmful (\emph{Denoising}) for imitating the expert and this determines whether features or learning from scratch performs best. Table~\ref{fig:universality}c shows that regardless of the which approach worked best, \sidetuning{} consistently matched or beat it.


\textbf{Reinforcement Learning for Navigation in Habitat.} Using a different learning algorithm (PPO) and direct interaction instead of expert trajectories, we observe identical trends. We trained agents directly in Habitat (74 buildings). Table~\ref{fig:universality}d shows performance in 14 held-out buildings after 10M frames of training. \Sidetuning{} performs comparably to the $\max$ of competing approaches.  

\subsection{Learning Mechanics in Side-Tuning}\label{sec:learning_mechanics}
\textbf{Using non-network base models.}
Since \sidetuning{} treats the base model is a black box, it can be used even when the base model is not a neural network. On \itaskonomy{}, we show that side-tuning can effectively use ground truth curvature as a base for incremental learning whereas all the methods we compare against cannot use this information (with the exception of feature extraction). 
Specifically, we resize the curvature image from $256{\times}256{\times}2$ to $32{\times}32{\times}2$ and reshape it to $16{\times}16{\times}8$, the same size as the output of other base models. 
Side-tuning with ground truth curvature achieves a better rank (4.3) on \itaskonomy{} than \textit{all 20} other methods (excluding Independent, 4.2) 



\textbf{Benefits for intermediate amounts of data.} 
We showed in the previous 
\begin{wrapfigure}{r}{0.45\textwidth}
    \vspace{-7mm}
    \centering
    \includegraphics[width=0.45\columnwidth]{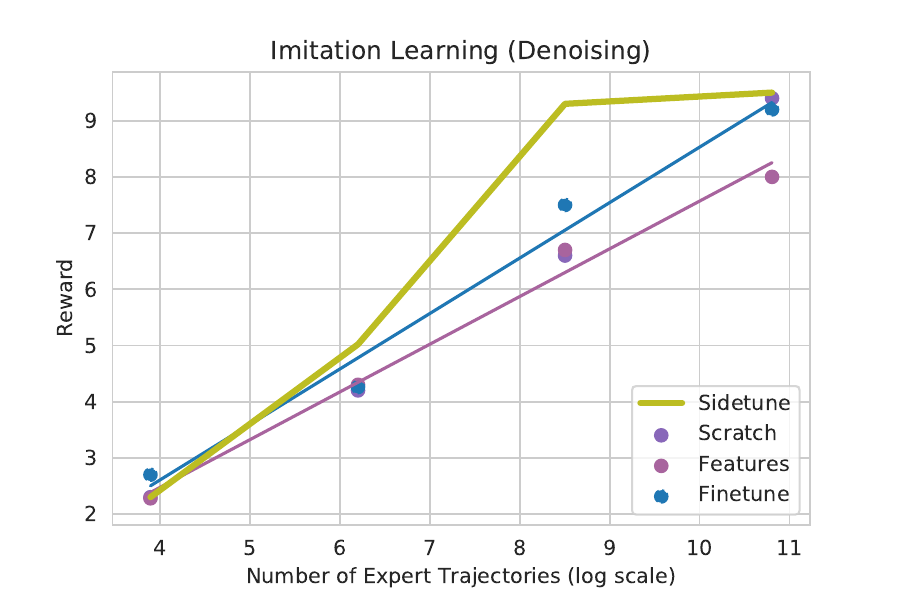}
    \vspace{-7mm}
    \caption{\footnotesize{\textbf{\Sidetuning{} outperformed features and fine-tuning on intermediate amounts of data.} Using imitation learning on a point-goal navigation task (setup from~\cite{habitat19iccv}).}}
    \label{fig:intermediate_data}
    \vspace{-6mm}
\end{wrapfigure}
section that \sidetuning{} performs like the best of $\{\text{features}, \text{fine-tuning}, \text{scratch}\}$ in domains with abundant or scant data. 

In order to test whether \sidetuning{} could profitably synthesize the features with \emph{intermediate} amounts of data, we evaluated each approach's ability to learn to navigate using 49, 490, 4900, or 49k expert trajectories and pretrained \emph{denoising} features. \Sidetuning{} was always the best-performing approach and, on intermediate amounts of data (e.g. 4.9k trajectories), outperformed the other techniques 
(\sidetune{} 9.3, fine-tune: 7.5, features: 6.7, scratch: 6.6), Figure~\ref{fig:intermediate_data}).

\begin{figure}
    \vspace{-2mm}
    \centering
    \includegraphics[width=0.46\columnwidth]{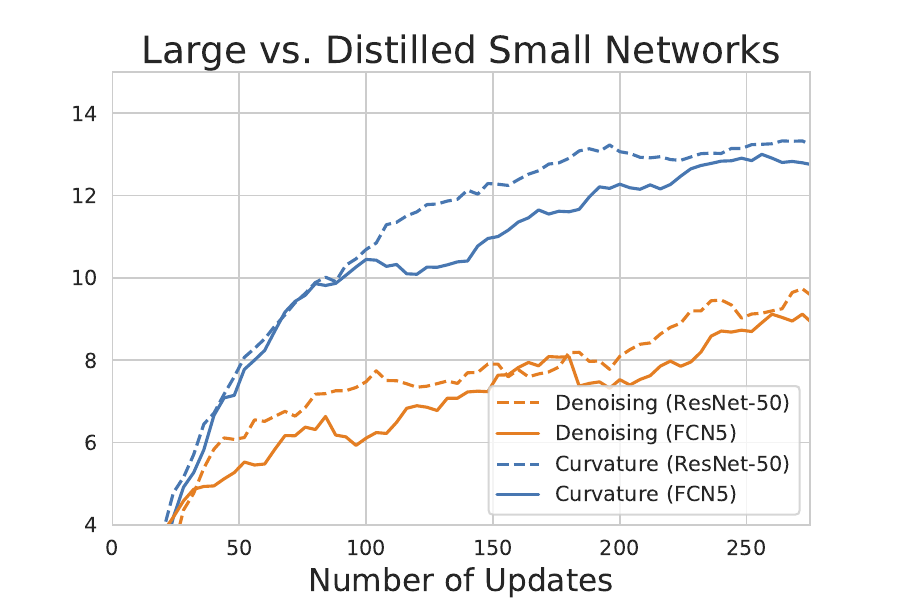}
    \includegraphics[width=0.39 \columnwidth]{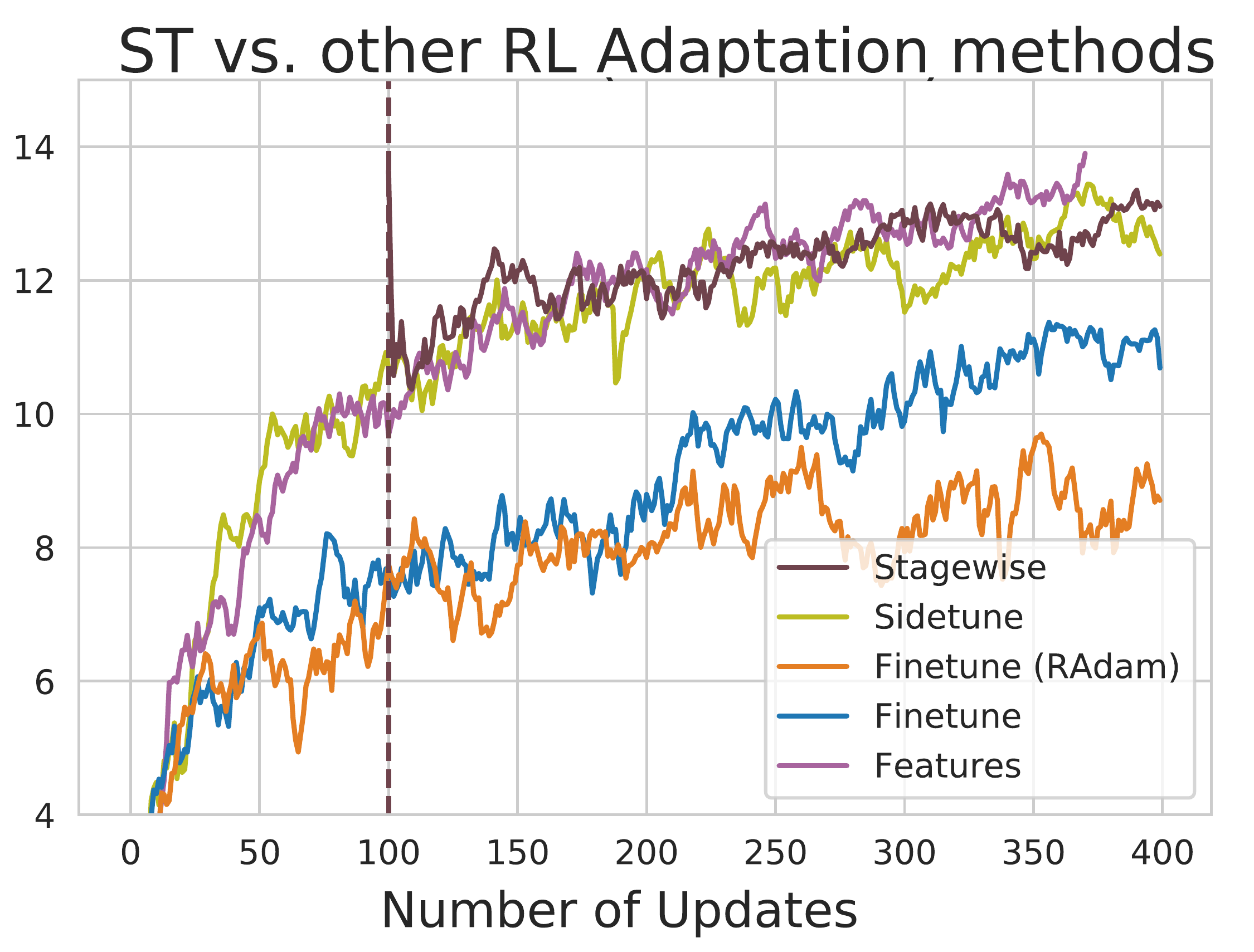}
    \vspace{-2mm}
    \caption{\textbf{Effect of network size in RL.} \textbf{(Left)} The distilled (5-layer) networks perform almost as well as the original ResNet-50 base networks. The choice of which base to use (Curvature vs. Denoising) has a much larger impact. \textbf{(Right)} Fine-tuning, even with RAdam, performed significantly worse than the alternative approaches.}
    \label{fig:rl_model_size}
    \vspace{-6mm}
\end{figure}

\textbf{Network size}. We find that when the base network is large, distilling it into a smaller network and \emph{sidetuning} will still retain most of the performance. In \figref{fig:rl_model_size}, we explore this in Habitat (RL using $\{\text{curvature}, \text{denoise}\} \rightarrow \text{navigation}$), with other results in the supplementary.

\textbf{More than just stable updates.}
In RL, fine-tuning often fails to improve performance. One common rationalization is that the early updates in RL are `high variance'. The common stage-wise solution is to first train using fixed features and then unfreeze the weights. We found that this approach performs as well (but no better than) using fixed features--and \sidetuning{} performed as well as both while not being domain-specific (Fig.~\ref{fig:rl_model_size}). We tested the `high-variance'
\begin{wrapfigure}{r}{0.45\textwidth}
    \centering
    \begin{minipage}{0.45\columnwidth}
        \centering
        \includegraphics[width=\columnwidth]{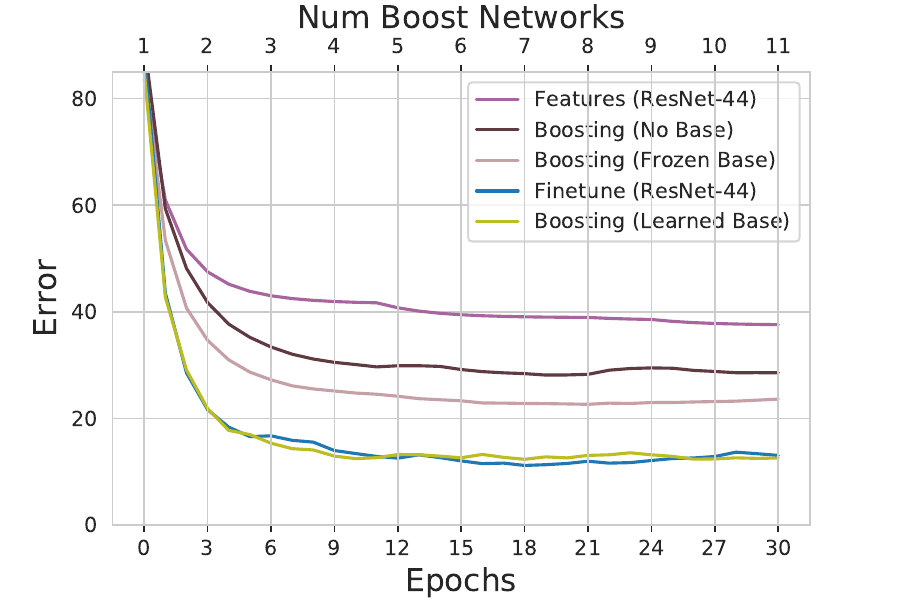}
        \vspace{0mm}
    \end{minipage}
    \vspace{-7mm}
    \caption{\footnotesize{\textbf{Boosting}. Deeper networks outperform many shallow learners.}}
    \vspace{-6mm}
\label{fig:boosting_and_radam}
\end{wrapfigure}
theory by fine-tuning with both gradient clipping and an optimizer designed to prevent such high-variance updates (RAdam~\cite{liu2019variance}). This provided no benefits over vanilla fine-tuning, suggesting that the benefits of side-tuning are not solely due to gradient stabilization early in training.

\textbf{Not Boosting.} Since the side network learns a residual on top of the base network, could \sidetuning{} be used for boosting? Although network boosting does improve performance on iCIFAR (Figure~\ref{fig:boosting_and_radam}), the parameters would've been better used in a deeper network rather than many shallow networks.

\subsection{Analysis of Design Choices}
We evaluate the effect of our architectural design decisions on task performance. 

\textbf{Base and Side Elements.}
\Sidetuning~uses two streams of information - one from the base model and one from the side model. Are both streams necessary? Table~\ref{fig:model-ablation} shows that on the \itaskonomy{} experiment performance improves when using both models. 

\textbf{Merge methods.}
\label{sec:merge_methods}
\secref{sec:alphablending} described different ways to merge the base and side networks. Table~\ref{fig:merge_methods} evaluates a few of these approaches. Product and alpha-blending are two of the simplest approaches and have little overhead in terms of compute and parameter count.
\cite{prognets16} (MLP) and \cite{perez2018film} (FiLM) use multi-layer perceptrons to adapt the base network to the new task. Table~\ref{fig:merge_methods} shows that alpha-blending, MLP, and FiLM are comparable, though the FiLM-based methods achieve marginally better average rank on \itaskonomy{}. We use alpha-blending as it adds fewer parameters and achieves similar performance.

    \begin{table*} 
        \vspace{-2mm}
        \scriptsize 
        \centering
        \begin{minipage}{0.50\textwidth}
            \centering
            \scriptsize
            \begin{tabular}{l|l}
        			\hline
        			\multicolumn{1}{|c|}{} & \multicolumn{1}{c|}{Avg. Rank ($\downarrow$)} \\ 
        			\multicolumn{1}{|c|}{Method} & \multicolumn{1}{c|}{\itaskonomy{}} \\
        			\hline
        
        			\hline
        			\multicolumn{1}{|c|}{{Product (Element-wise)}} & \multicolumn{1}{c|}{3.64} \\
        			\multicolumn{1}{|c|}{{\textit{Summation ($\alpha$-blending)}}} & \multicolumn{1}{c|}{2.27} \\
        			\multicolumn{1}{|c|}{{MLP (\cite{prognets16})}} & \multicolumn{1}{c|}{2.18} \\
        			\multicolumn{1}{|c|}{{FiLM \cite{perez2018film}}} & \multicolumn{1}{c|}{1.91} \\
        			\hline
        	\end{tabular}
        	\vspace{2mm}
            \captionof{table}{\footnotesize{\textbf{Average rank of various merge methods.} Alternative feature-wise transformations did not outperform simple $\alpha$-blending in a statistically significant way. }}  \label{fig:merge_methods}        
        \end{minipage}
        \hspace{2mm}
        \begin{minipage}{0.42\textwidth}
            \centering
            \scriptsize
            \begin{tabular}{l|l}
        			\hline
        			\multicolumn{1}{|c|}{} & \multicolumn{1}{c|}{Avg. Rank ($\downarrow$)} \\ 
        			\multicolumn{1}{|c|}{Method} & \multicolumn{1}{c|}{\itaskonomy{}} \\
        			\hline
        
        			\hline
        			\multicolumn{1}{|c|}{{Base-Only}} & \multicolumn{1}{c|}{2.55} \\
        			\multicolumn{1}{|c|}{{Side-Only}} & \multicolumn{1}{c|}{2.10} \\
        			\multicolumn{1}{|c|}{{\Sidetuning{}}} & \multicolumn{1}{c|}{1.36} \\
        			\hline
        	\end{tabular}
        	\vspace{2mm}
            \caption{\footnotesize
            {\textbf{Both base and side networks contribute.} Performance (average rank on \itaskonomy{}) improved when using \textit{both} the base and side model in \sidetuning.}} \label{fig:model-ablation}
        \end{minipage}
        \vspace{-6mm}
        \addtolength{\tabcolsep}{2pt}
    \end{table*}
    
\textbf{Side Network Initialization.} A good side network initialization can yield a minor boost in performance. We found that initializing from the base network slightly outperforms a low-energy initialization\footnote{Side network is trained to not impact the output. Full details in the supplementary.}, which slightly outperforms Xavier initialization. However, we found that these differences were not statistically significant across tasks ($H_0: \textit{pretrained} = \textit{xavier}; p=0.07$, Wilcoxon signed-rank test). We suspect that initialization might be more important on harder problems. We test this by repeating the analysis without the simple texture-based tasks (2D keypoint + edge detection and autoencoding) and find the difference in initialization is now significant ($p=0.01$).

\section{Conclusions and Limitations}
We have introduced the \sidetuning{} framework, a simple yet effective approach for \emph{additive learning}. Since it does not suffer from catastrophic forgetting or rigidity, it is naturally suited to incremental learning. The theoretical advantages are reflected in empirical results, and we found \sidetuning{} to perform on par with \textit{or better than} many current incremental learning approaches, while being significantly simpler. Experiments demonstrated this in challenging contexts and with various state-of-the-art neural networks across multiple domains. 

More complex methods should need to demonstrate clear improvements over simply doing this na\"ive approach. We see several natural ways to improve it:

\textit{Better forward transfer}: Our experiments used only a single base and single side network. Leveraging previously trained side networks could yield better performance on later tasks.

\textit{Learning when to deploy side networks}: Like most incremental learning setups, we assumed that the tasks are presented in a sequence and that task identities are known. 
Using several active side networks in tandem would provide a natural way to identify task change or distribution shift.

\textit{Using side-tuning to measure task relevance}: We found that $\alpha$ tracked task relevance in~\cite{taskonomy2018}, but a more rigorous treatment of the interaction between the base, side, $\alpha$ and final performance could yield insight into how tasks relate.

\vspace{3mm}
\noindent\footnotesize{\textbf{Acknowledgements:} This material is based upon work supported by ONR MURI (N00014-14-1-0671), Vannevar Bush Faculty Fellowship, an Amazon AWS Machine Learning Award, NSF (IIS-1763268), a BDD grant and TRI. Toyota Research Institute (``TRI'') provided funds to assist the authors with research but this article solely reflects the opinions and conclusions of its authors and not TRI or any other Toyota entity.}

\medskip

\clearpage
%
%
\bibliographystyle{splncs04}
\bibliography{main_submission}
\end{document}


\pagestyle{headings}
\mainmatter
\def\ECCVSubNumber{1104}
\title{Side-Tuning: A Baseline for Network Adaptation via Additive Side Networks: Supplementary Material}
\titlerunning{Side-Tuning: Supplementary Material}
\author{
    Jeffrey O. Zhang \inst{1} \and
    Alexander Sax \inst{1} \and
    Amir Zamir \inst{1,2} \and
    Leonidas Guibas \inst{2} \and
    Jitendra Malik \inst{1}
}
\authorrunning{Zhang et al.}
\institute{UC Berkeley \and Stanford University \\ 	\textcolor{blue}{\url{http://sidetuning.berkeley.edu}\vspace{-9pt}}}

\maketitle
\begin{abstract}
The following items are provided in the supplementary material:
\begin{enumerate}
    \item Experimental Details 
    \item Additional Experiments
    \item iCIFAR Data
    \item iTaskonomy Data
\end{enumerate}
\end{abstract}

\section{Experimental Details} \label{sec:experimental_setup}
For full details on our configs, please refer to ./configs in provided code on our website at \href{http://sidetuning.berkeley.edu}{http://sidetuning.berkeley.edu}. 

\subsection{Experimental Setup for Incremental Learning} 

\subsubsection{Taskonomy} Our data is 4M images on 12 single image tasks. The tasks that we use are the following: curvature, semantic segmentation, reshading, keypoints3d, keypoints2d, texture edges, occlusion edges, distance, depth, surface normals, object classification and autoencoding. The tasks were chosen in no particular special order. Our base model and side model are ResNet-50s. 
We pretrain on curvature. Then, we train each task for three epochs before moving on to the next task. We use cross entropy loss for classification tasks (semantic segmentation and object classification), L2 loss for curvature and L1 loss for the other tasks. We use Adam optimizer with an initial learning rate of 1e-4, weight decay coefficient of 2e-6, gradient clipping to 1.0, and batch size of 32.
We evaluate our performance on a held out set of images, both immediately after training a specific task, and after training of all the tasks are complete. 
    
\subsubsection{iCIFAR} We start by pretraining a model on CIFAR 10 (from \url{https://github.com/akamaster/pytorch_resnet_cifar10}). Then we partition CIFAR100 into 10 distinct sets of 10 classes. Then, we train for 4 epochs on these tasks using Adam optimizer, learning rate of 1e-3, batch size of 128.
    
\subsection{Experimental Setup for Additional Domains} 
\subsubsection{NLP} We train and test on the the question answering dataset SQuAD2.0, a reading comprehension dataset consisting of 100,000 questions with 50,000 unanswerable questions. Both our base encoding and side network is a BERT transformer pretrained on a larger corpus. Finetuning trains a single BERT transfer. We use the training setup found at \url{https://github.com/huggingface/pytorch-transformers} (train for 2 epochs at a learning rate of 3e-5) with one caveat - we use an effective batch size of 3 (vs. their 24).
    
\subsubsection{Habitat Experiments} 
We borrow the experimental setup from \cite{midLevelReps2019}:
    
    \begin{quote}
        We use the Habitat environment with the Gibson dataset. The dataset virtualizes 572 actual buildings, reproducing the intrinsic visual and semantic complexity of real-world scenes. 
    
        We train and test our agents in two disjoint sets of buildings. During testing we use buildings that are different and completely unseen during training.  We use up to 72 building for training and 14 test buildings for testing.  The train and test spaces comprise 15678.4$m^2$ (square meters) and 1752.4$m^2$, respectively.
    
        The agent must direct itself to a given nonvisual target destination (specified using coordinates), avoiding obstacles and walls as it navigates to the target. The maximum episode length is 500 timesteps, and the target distance is between 1.4 and 15 meters from the start. 
    \end{quote}
    This setup is shared between imitation learning and RL, which differ in the data, architecture and optimization process. 
    
    \textbf{Imitation Learning}
    We collect 49,325 shortest path expert trajectories in Habitat, 2,813,750 state action pairs. We learn a neural network mapping from states to actions. Our base encoding is a ResNet-50 and the side network is a five layer convolutional network. The representation output is then fed into a neural network policy. We train the model for 10 epochs using cross entropy loss and Adam at an initial learning rate of 2e-4 and weight decay coefficient of 3.8e-7. We initialize alpha to 0.5. Finetuning uses the same model architecture but updates all the weights. Feature extraction only uses the ResNet-50 to collect features. 
    
    \textbf{RL}
    Similarly, we borrow the RL setup from \cite{midLevelReps2019}.
    \begin{quote}
        In all experiments we use the common Proximal Policy Optimization (PPO) algorithm with Generalized Advan- tage Estimation. Due to the computational load of ren- dering perceptually realistic images in Gibson we are only able to use a single rollout worker and we therefore decorre- late our batches using experience replay and off-policy vari- ant of PPO. The formulation is similar to Actor-Critic with Experience Replay (ACER) in that full trajectories are sampled from the replay buffer and reweighted using the first-order approximation for importance sampling.
    \end{quote}
    During training, the agent receives a large one-time reward for reaching the goal, a positive reward proportional to Euclidean distance toward the goal and a small negative reward each timestep. The maximum episode length is 500 timesteps, and the target distance is between 1.4 and 15 meters from the start.

    Due to this paradigms' compute and memory constraints, it would be difficult for us to use large architectures for this setting. Thus, our base encoding is a five layer convolutional network distilled from the trained ResNet-50. Our side network is also a five layer convoultional network. Finetuning is handled the same way - update all the weights in this setup. Feature extraction uses the five layer network to collect features. 
    
\subsection{Experimental Setup for Learning Mechanics}
\textbf{Low energy initialization} In classical teacher student distillation, the student is trained to minimize the distance between its output and the teacher's output. In this setting, we minimize the distance between the teacher's output and the summation of the student's output and the teacher's output. The output space may have a different geometry than that of the input space. 

\section{Additional Experiments}
\subsection{Task relevance predicts alpha $\alpha$.} In our experiments, we treat $\alpha$ as a learnable parameter (initialized to 0.5) and find that the relative values of $\alpha$ are predictive of empirical performance. In imitation learning, \emph{curvature} ($\alpha=0.557$) outperformed \emph{denoising} ($\alpha=0.252$). In \itaskonomy{}, the $\alpha$ values from training on just 100 images predicted the actual transfer performance to normals in~\cite{taskonomy2018}, (e.g. \emph{curvature} ($\alpha=0.56$) outperformed \emph{object classification} ($\alpha=0.50$)). For small datasets, usually $\alpha \approx 0.5$ and the relative order, rather than the actual value is important.

\subsection{Fusion}
An alternative perspective views these methods as various fusions between some base output and new side output. In this framework, side-tuning is a late-fusion approach whereas PNN is a distributed-fusion approach. 
We compare various fusion methods in iCIFAR and find that late fusion performs better than early fusion and distributed fusion (error of 23.9 vs. 38.8 and 26.3 respectively). The fusion is performed with a MLP. 
We run this analysis in \itaskonomy{} as well. We fuse with no new parameters, merging with summation and combine outputs with alpha-blending. We find that late fusion outperforms early and distributed fusion as well (rank of 1.25 of 3 vs. 1.92 and 2.0 respectively). 

\subsection{Imitation Learning Data Study}
We show additional results from the data study in imitaiton learning in \figref{fig:il_data_more}. The results overall show \Sidetuning{}'s advantage.
\begin{figure}
    \vspace{-5mm}
    \begin{minipage}{0.3\textwidth}
        \centering
        \includegraphics[width=\textwidth]{supmat_figures/il_data_denoising_spl-2.pdf}
        \vspace{0mm}
    \end{minipage}
    \begin{minipage}{0.3\textwidth}
        \centering
        \includegraphics[width=\textwidth]{supmat_figures/il_data_curvature_spl-2.pdf}
        \vspace{0mm}
    \end{minipage}
    \begin{minipage}{0.3\textwidth}
        \centering
        \includegraphics[width=\textwidth]{supmat_figures/il_data_curvature_reward-2.pdf}
        \vspace{0mm}
    \end{minipage}
    \caption{Additional Imitation Learning Data Study. We ablate over different quantities of expert trajectories. We observe that when data is scarce, features is a powerful choice whereas when data is plentiful, fine-tuning performs well. In both scenarios, \sidetuning{} is able to perform as well as the stronger approach.}
    \label{fig:il_data_more}
\end{figure}

\subsection{Additional iCIFAR Comparisons with Progressive Neural Networks}
In the main paper, for our iCIFAR experiments, we see that the average performance of side-tuning weaker than that of PNN. We find that side-tuning can bridge this gap with a multilayer perceptron (adapter) to merge the base and side networks. This is a common practice in PNN. We see with the adapter network, the two methods are very similar when measuring classification error (23.69 of PNN vs. 23.91 of \Sidetuning{}). 

\subsection{Extremely Few-Shot Learning}
In domains with very few examples, we found that \sidetuning{} is unable to match the performance of other methods. We evaluated our setup in vision transfer for 5 images from the same building, imitation learning given 5 expert trajectories.

\vspace{5mm}
\begin{tabular}{ll}
        \hline
         \multicolumn{1}{|c|}{} \\
         \multicolumn{1}{|c|}{Methods} \\
        \hline
        \hline
        \multicolumn{1}{|c|}{Finetune} \\
         \multicolumn{1}{|c|}{Features} \\
        \multicolumn{1}{|c|}{Scratch} \\
        \hline
        \multicolumn{1}{|c|}{Sidetune}\\
       \hline
\end{tabular}
\begin{tabular}{ll}
	\hline
	 \multicolumn{2}{|c|}{Nav. Rew. (4 epi) ($\uparrow$)} \\ 
	 \multicolumn{1}{|c}{\scriptsize{Curvature}} & \multicolumn{1}{c|}{\scriptsize{Denoise}} \\
	\hline
	\hline
	\multicolumn{1}{|c}{-0.1} & \multicolumn{1}{c|}{-1.2} \\
	 \multicolumn{1}{|c}{0.4} & \multicolumn{1}{c|}{1.2} \\
	\multicolumn{1}{|c}{-0.9} & \multicolumn{1}{c|}{-0.9} \\
	\hline
	\multicolumn{1}{|c}{-0.3} & \multicolumn{1}{c|}{-1.8} \\
	\hline
\end{tabular}
\begin{tabular}{l}
        \hline
         \multicolumn{1}{|c|}{Loss (5 ims) ($\downarrow$)} \\ 
         \multicolumn{1}{|c|}{\scriptsize{Curvature} to Normals} \\
        \hline
        \hline
        \multicolumn{1}{|c|}{0.35} \\
         \multicolumn{1}{|c|}{0.36} \\
        \multicolumn{1}{|c|}{0.37} \\
        \hline
        \multicolumn{1}{|c|}{0.42}\\
       \hline
\end{tabular}

\newpage
\section{iCIFAR data}
\subsection{Average Final Classification error}
Here, we show the average classification error at the end of training. 

\begin{small}
\begin{tabular}{|l|r|}
\hline
{} &      Avg. Error \\
\hline
Indep.            &  14.81 \\
PNN               &  23.69 \\
Side-tuning (MLP) &  23.91 \\
Late Fusion       &  23.91 \\
Dist. Fusion      &  26.34 \\
Side-tuning       &  32.77 \\
EWC               &  34.86 \\
Early Fusion      &  38.84 \\
EWC $\lambda=10^4$&  41.39 \\
EWC $\lambda=10^6$&  41.88 \\
Feat.             &  42.27 \\
Res. Adapter      &  46.26 \\
Piggyback         &  47.72 \\
PSP               &  50.82 \\
Fine-tuning       &  77.75 \\
\hline
\end{tabular}
\end{small}

\subsection{Final Classification error for all methods on all tasks}
\begin{small}
\begin{tabular}{|l|r|r|r|r|r|r|r|r|r|r|}
\hline
{} &  Task 0 &  Task 1 &  Task 2 &  Task 3 &  Task 4 &  Task 5 &  Task 6 &  Task 7 &  Task 8 &  Task 9 \\
\hline
Side-tuning       &    31.7 &    31.2 &    33.1 &    36.4 &    30.3 &    34.6 &    33.0 &    36.8 &    33.1 &    27.5 \\
PSP               &    57.9 &    63.9 &    50.9 &    60.4 &    47.0 &    51.5 &    46.6 &    47.5 &    50.3 &    32.2 \\
Fine-tuning       &    94.2 &    87.3 &    91.9 &    91.0 &    89.3 &    88.9 &    88.0 &    79.3 &    54.6 &    13.0 \\
Res. Adapter      &    47.2 &    53.7 &    46.8 &    49.1 &    45.5 &    43.3 &    45.1 &    43.2 &    44.5 &    44.2 \\
Side-tuning (MLP) &    26.2 &    24.1 &    23.4 &    26.9 &    23.3 &    23.8 &    23.0 &    26.1 &    23.8 &    18.5 \\
PNN               &    25.4 &    27.2 &    24.6 &    23.9 &    23.2 &    24.5 &    22.4 &    28.6 &    22.0 &    15.1 \\
Indep.            &    12.2 &    17.2 &    12.8 &    13.8 &    16.1 &    17.1 &    15.9 &    18.5 &    13.5 &    11.0 \\
EWC               &    19.6 &    33.1 &    35.6 &    41.6 &    35.6 &    41.4 &    37.1 &    38.7 &    40.1 &    25.8 \\
Piggyback         &    48.7 &    48.0 &    45.9 &    51.0 &    49.2 &    48.8 &    46.1 &    49.9 &    43.6 &    46.0 \\
Feat.             &    45.3 &    37.9 &    47.1 &    41.0 &    40.1 &    45.5 &    42.8 &    46.3 &    40.9 &    35.8 \\
EWC $\lamda=10^6$              &    14.1 &    41.1 &    45.0 &    47.6 &    42.2 &    44.5 &    46.3 &    47.0 &    47.2 &    38.9 \\
EWC $\lamda=10^4$              &    56.5 &    54.5 &    50.6 &    45.0 &    42.7 &    39.5 &    40.0 &    39.1 &    34.7 &    16.2 \\
Early Fusion      &    38.4 &    44.9 &    39.4 &    43.6 &    37.8 &    36.5 &    41.0 &    36.9 &    37.8 &    32.1 \\
Late Fusion       &    26.2 &    24.1 &    23.4 &    26.9 &    23.3 &    23.8 &    23.0 &    26.1 &    23.8 &    18.5 \\
Dist. Fusion      &    26.1 &    27.7 &    26.6 &    27.7 &    23.9 &    29.4 &    27.1 &    27.9 &    25.8 &    21.2 \\
\hline
\end{tabular}
\end{small}

\newpage \section{iTaskonomy Data}
\label{sec:iTaskonomy}
\subsection{Rank of \Sidetuning{} and Baselines}
Here we provide rankings for the baselines introduced.

\begin{small}
\begin{tabular}{|l|r|}
\hline
{} &      Rank \\
\hline
Side-tuning  &  2.000 \\
Indep.       &  2.000 \\
PNN          &  3.500 \\
Piggyback    &  4.333 \\
Feat.        &  5.000 \\
Res. Adapter &  5.250 \\
EWC          &  7.583 \\
PSP          &  7.667 \\
Fine-tuning  &  7.667 \\
\hline
\end{tabular}
\end{small}

\subsection{Rank of all methods}
Here, we provide the rank of all methods that we have tested in the incremental learning setup (including all ablations/analysis in addition to baselines). Note the effectiveness of using ground truth curvatures as the base model!
\begin{small}
\begin{tabular}{|l|r|}
\hline
{} &       Rank \\
\hline
Indep.                 &   4.167 \\
GT Curv as Base        &   4.333 \\
Side-tuning            &   5.250 \\
FiLM                   &   5.667 \\
MLP                    &   6.417 \\
Dist. Fusion           &   7.000 \\
Xavier Init.           &   8.500 \\
MLP2                   &   8.583 \\
PNN                    &   8.833 \\
Low Energy Init.       &   9.167 \\
No Base                &  10.583 \\
Mult.                  &  10.583 \\
Piggyback              &  10.750 \\
Early Fusion           &  11.667 \\
Feat.                  &  12.417 \\
No Side                &  12.417 \\
Res. Adapter           &  14.167 \\
EWC                    &  17.667 \\
$EWC (\lambda=0.01)$ &  17.750 \\
$EWC (\lambda=100)$  &  18.000 \\
Fine-tuning            &  18.000 \\
PSP                    &  18.583 \\
\hline
\end{tabular}
\end{small}

\newpage

\subsection{Final Task Performance for all Tasks and Methods}
Here we provide the final performance for all of the tested methods on all the incremental learning tasks in Taskonomy. 

\begin{figure}
\begin{small}
\begin{tabular}{|l|r|r|r|r|r|r|r|}
\hline
{} &  Side-tuning &     PSP &  Fine-tuning &    PNN &  Indep. &    EWC &  Feat. \\
\hline
Curvature             &        1.234 &   1.810 &        1.787 &  1.230 &   1.251 &  1.829 &  1.240 \\
Semantic Segmentation &        1.347 &   1.626 &        1.505 &  1.350 &   1.347 &  1.502 &  1.353 \\
Reshading             &        0.613 &   2.377 &        2.127 &  0.621 &   0.638 &  1.677 &  0.791 \\
3D Keypoints          &        1.986 &   3.026 &        3.486 &  2.038 &   1.993 &  3.096 &  1.988 \\
2D Keypoints          &        0.896 &   4.593 &        5.046 &  1.042 &   0.742 &  4.988 &  2.504 \\
Texture Edges         &        0.160 &   1.571 &        1.374 &  0.165 &   0.141 &  1.007 &  0.377 \\
Occlusion Edges       &        0.927 &   1.147 &        1.177 &  0.936 &   0.933 &  1.235 &  0.928 \\
Z-buffer Depth        &        1.011 &   3.222 &        3.587 &  1.060 &   1.018 &  3.706 &  1.214 \\
Distance              &        1.117 &   2.900 &        3.401 &  1.098 &   1.001 &  3.860 &  1.187 \\
Surface Normals       &        0.618 &   1.912 &        2.040 &  0.644 &   0.569 &  1.990 &  0.649 \\
Object Classification &        2.986 &  15.950 &        5.042 &  2.996 &   2.829 &  5.103 &  3.183 \\
Autoencoding          &        1.271 &   1.337 &        1.179 &  1.477 &   1.138 &  1.155 &  5.337 \\
\hline
\end{tabular}
\end{small}\\

\vspace{10mm}

\begin{small}
\begin{tabular}{|l|r|r|r|r|r|r|}
\hline
{} &  Piggyback &  Res. Adapter &  Low E Init. &  Xavier Init. &  No Base &  No Side \\
\hline
Curvature             &      1.245 &         1.370 &             1.239 &         1.237 &    1.294 &    1.240 \\
Semantic Segmentation &      1.350 &         1.382 &             1.351 &         1.352 &    1.353 &    1.353 \\
Reshading             &      0.646 &         0.743 &             0.699 &         0.702 &    0.923 &    0.791 \\
3D Keypoints          &      1.976 &         2.226 &             1.984 &         1.988 &    2.060 &    1.988 \\
2D Keypoints          &      2.442 &         1.009 &             0.890 &         0.889 &    0.870 &    2.504 \\
Texture Edges         &      0.355 &         0.171 &             0.154 &         0.160 &    0.155 &    0.377 \\
Occlusion Edges       &      0.935 &         0.979 &             0.929 &         0.930 &    0.944 &    0.928 \\
Z-buffer Depth        &      1.120 &         1.314 &             1.264 &         1.163 &    1.045 &    1.214 \\
Distance              &      1.170 &         1.387 &             1.179 &         1.138 &    1.219 &    1.187 \\
Surface Normals       &      0.638 &         0.723 &             0.645 &         0.638 &    0.651 &    0.649 \\
Object Classification &      3.187 &         3.044 &             3.126 &         3.099 &    2.960 &    3.183 \\
Autoencoding          &      4.939 &         1.298 &             1.312 &         1.256 &    1.262 &    5.337 \\
\hline
\end{tabular}
\end{small}
\end{figure}

\begin{figure}
\begin{small}
\begin{tabular}{|l|r|r|r|r|r|}
\hline
{} &  GT Curv as Base &  $EWC (\lambda=0.01)$ &  $EWC (\lambda=100)$ &   MLP2 &    MLP \\
\hline
Curvature             &            0.783 &                 1.626 &                1.905 &  1.244 &  1.247 \\
Semantic Segmentation &            1.351 &                 1.472 &                1.496 &  1.351 &  1.351 \\
Reshading             &            0.591 &                 1.955 &                1.668 &  0.603 &  0.844 \\
3D Keypoints          &            1.702 &                 3.125 &                3.229 &  2.000 &  1.997 \\
2D Keypoints          &            0.879 &                 4.649 &                6.445 &  0.967 &  0.865 \\
Texture Edges         &            0.151 &                 1.036 &                0.969 &  0.164 &  0.150 \\
Occlusion Edges       &            0.928 &                 1.191 &                1.320 &  0.932 &  0.926 \\
Z-buffer Depth        &            1.050 &                 3.726 &                3.168 &  1.074 &  1.011 \\
Distance              &            1.076 &                 3.640 &                3.389 &  1.132 &  1.080 \\
Surface Normals       &            0.562 &                 2.139 &                2.112 &  0.619 &  0.612 \\
Object Classification &            2.956 &                 5.280 &                5.680 &  2.979 &  2.955 \\
Autoencoding          &            1.294 &                 1.163 &                1.223 &  1.383 &  1.312 \\
\hline
\end{tabular}
\end{small}

\vspace{10mm}

\begin{small}
\begin{tabular}{|l|r|r|r|r|}
\hline
{} &  Mult. &   FiLM &  Early Fusion &  Dist. Fusion \\
\hline
Curvature             &  1.235 &  1.243 &         1.318 &         1.230 \\
Semantic Segmentation &  1.354 &  1.352 &         1.364 &         1.348 \\
Reshading             &  0.813 &  0.827 &         0.720 &         0.631 \\
3D Keypoints          &  1.990 &  1.979 &         2.148 &         1.984 \\
2D Keypoints          &  1.369 &  0.840 &         0.855 &         0.974 \\
Texture Edges         &  0.224 &  0.145 &         0.166 &         0.167 \\
Occlusion Edges       &  0.934 &  0.928 &         0.948 &         0.925 \\
Z-buffer Depth        &  1.097 &  1.028 &         1.158 &         1.048 \\
Distance              &  1.085 &  1.075 &         1.219 &         1.083 \\
Surface Normals       &  0.625 &  0.610 &         0.691 &         0.626 \\
Object Classification &  3.005 &  2.991 &         3.022 &         2.999 \\
Autoencoding          &  2.012 &  1.264 &         1.224 &         1.275 \\
\hline
\end{tabular}
\end{small}
\end{figure}

\newpage
\subsection{Qualitative Results of \Sidetuning{} vs PNN}
We show predictions for \Sidetuning{} and PNN side-by-side for three tasks (relative to Ground Truth and Independent). We show two variants of PNN, the first following the paper closely, the second with minor variations. 

\begin{figure}
    \vspace{-8mm}
    \centering
    \includegraphics[width=0.75\textwidth]{supmat_figures/main_2_labeled.pdf}
    \caption{\footnotesize{\textbf{Qualitative results for Reshading.} Both PNN methods and Sidetune have similar qualitative results.  }}
    \label{fig:vs_pnn_qual_2}
\end{figure}

\begin{figure*}
    \centering
    \includegraphics[width=0.75\textwidth]{supmat_figures/main_5_labeled.pdf}
    \caption{\footnotesize{\textbf{Qualitative results for 2D Edges.} Both PNN methods and Sidetune have similar qualitative results. }}
    \label{fig:vs_pnn_qual_5}
\end{figure*}

\begin{figure*}
    \centering
    \includegraphics[width=0.75\textwidth]{supmat_figures/main_9_labeled.pdf}
    \caption{\footnotesize{\textbf{Qualitative results for Surface Normals.} Both PNN methods and Sidetune have similar qualitative results. }}
    \label{fig:vs_pnn_qual_9}
\end{figure*}

\subsection{Qualitative Results of Baselines}\label{sec:qualitative_lll}
We show predictions for each method (\Sidetuning{}, EWC, PSP, PNN, Independent) separately for some fixed set of randomly selected images throughout training.

\begin{figure*}[h]
    \centering
    \textbf{Side-Tuning}
    \newline
    \includegraphics[width=0.8\textwidth]{supmat_figures/sidetune.pdf}
    \caption{\footnotesize{\textbf{More qualitative results for \sidetuning{}.} These images were randomly selected from the validation set. Left-hand column is input, rightmost-column is ground truth. Images from left to right show predictions as training progresses. Each block of 4 rows shows predictions on a different task (\emph{Reshading}, \emph{2D Edges}. \emph{Surface Normals}.)}}
    \vspace{25mm}
\end{figure*}

\begin{figure*}
    \centering
    \textbf{Elastic Weight Consolidation (EWC)}
    \includegraphics[width=\textwidth]{supmat_figures/ewc.pdf}
    \caption{\footnotesize{\textbf{More qualitative results for EWC.} These images were randomly selected from the validation set. Left-hand column is input, rightmost-column is ground truth. Images from left to right show predictions as training progresses. Each block of 4 rows shows predictions on a different task (\emph{Reshading}, \emph{2D Edges}. \emph{Surface Normals}.)}}
    \vspace{25mm}
\end{figure*}

\begin{figure*}
    \centering
    \textbf{Parameter Superposition (PSP)}
    \includegraphics[width=\textwidth]{supmat_figures/psp.pdf}
    \caption{\footnotesize{\textbf{More qualitative results for PSP.} These images were randomly selected from the validation set. Left-hand column is input, rightmost-column is ground truth. Images from left to right show predictions as training progresses. Each block of 4 rows shows predictions on a different task (\emph{Reshading}, \emph{2D Edges}. \emph{Surface Normals}.)}}
    \vspace{15mm}
\end{figure*}

\begin{figure*}
    \centering
    \textbf{Progressive Neural Network (PNN)}
    \includegraphics[width=\textwidth]{supmat_figures/pnn.pdf}
    \caption{\footnotesize{\textbf{More qualitative results for PNN.} These images were randomly selected from the validation set. Left-hand column is input, rightmost-column is ground truth. Images from left to right show predictions as training progresses. Each block of 4 rows shows predictions on a different task (\emph{Reshading}, \emph{2D Edges}. \emph{Surface Normals}.)}}
    \vspace{25mm}
\end{figure*}

\begin{figure*}
    \centering
    \textbf{Independent Fine-Tuned ResNet-50}
    \includegraphics[width=\textwidth]{supmat_figures/independent.pdf}
    \caption{\footnotesize{\textbf{More qualitative results for \emph{independent}} These images were randomly selected from the validation set. Left-hand column is input, rightmost-column is ground truth. Images from left to right show predictions as training progresses. Each block of 4 rows shows predictions on a different task (\emph{Reshading}, \emph{2D Edges}. \emph{Surface Normals}.)}}
    \vspace{15mm}
\end{figure*}

\begin{figure*}
    \centering
    \textbf{Piggyback}
    \includegraphics[width=\textwidth]{supmat_figures/piggyback.png}
    \caption{\footnotesize{\textbf{More qualitative results for \emph{piggyback}} These images were randomly selected from the validation set. Left-hand column is input, rightmost-column is ground truth. Images from left to right show predictions as training progresses. Each block of 4 rows shows predictions on a different task (\emph{Reshading}, \emph{2D Edges}. \emph{Surface Normals}.)}}
    \vspace{15mm}
\end{figure*}

\begin{figure*}
    \centering
    \textbf{Residual Adapter}
    \includegraphics[width=\textwidth]{supmat_figures/radapter.png}
    \caption{\footnotesize{\textbf{More qualitative results for \emph{Residual Adapter}} These images were randomly selected from the validation set. Left-hand column is input, rightmost-column is ground truth. Images from left to right show predictions as training progresses. Each block of 4 rows shows predictions on a different task (\emph{Reshading}, \emph{2D Edges}. \emph{Surface Normals}.)}}
    \vspace{15mm}
\end{figure*}

\clearpage
%
%
\bibliographystyle{splncs04}
\bibliography{main_submission}